\documentclass[11pt,a4paper]{article}
\usepackage{times,latexsym}
\usepackage{url}
\usepackage[T1]{fontenc}

\usepackage[acceptedWithA]{tacl2021v1}

\usepackage{amsmath}
\usepackage{bbold}
\usepackage{layouts}
\newcommand\thefont{\expandafter\string\the\font}

\usepackage{color-edits}
\addauthor[JZ]{jelle}{orange}
\addauthor[JJ]{jaap}{green}
\addauthor[JS]{jakub}{red}
\addauthor[LB]{lisa}{blue}  

\usepackage{tipa}
\usepackage{booktabs}
\usepackage{scalerel,xparse}
\usepackage{xspace} 
\usepackage{colortbl}
\usepackage{arydshln}
\usepackage{comment}
\usepackage{enumitem}

\newcommand\aopDelta{\mbox{AOP-$\Delta$}\xspace}

\definecolor{lr}{rgb}{1.0, 0.596078431372549, 0.5882352941176471}
\definecolor{dr}{rgb}{0.8392156862745098, 0.15294117647058825, 0.1568627450980392}
\definecolor{lg}{rgb}{0.596078431372549, 0.8745098039215686, 0.5411764705882353}
\definecolor{dg}{rgb}{0.17254901960784313, 0.6274509803921569, 0.17254901960784313}

\usepackage{xspace,mfirstuc,tabulary}

\newif\iftaclinstructions
\taclinstructionsfalse 
\iftaclinstructions
\renewcommand{\confidential}{}
\renewcommand{\anonsubtext}{(No author info supplied here, for consistency with
TACL-submission anonymization requirements)}
\newcommand{\instr}
\fi

\iftaclpubformat 

\else

\fi

\newcommand\Mark[1]{\textsuperscript{#1}}
\newcommand\new[1]{#1}

\newcommand\newest[1]{#1}

\title{Black Big Boxes: Do Language Models Hide a Theory of Adjective Order?}
\title{Black Big Boxes:\\ Do Language Models Acquire Soft Word Order Constraints?}
\title{Black Big Boxes: A case study on adjective order to explore the predictive and explanatory potential of large language models for linguistic theory}  
\title{Black Big Boxes:\\ Tracing Adjective Order Preferences in Large Language Models}

\author{Jaap Jumelet\Mark{1,3}~~~~~~~~Lisa Bylinina\Mark{2}~~~~~~~~Willem Zuidema\Mark{3}~~~~~~~~Jakub Szymanik\Mark{4} \\[5pt]
\begin{tabular}{c c c c} 
\small\Mark{1}CLCG & \small\Mark{2}ILS & \small\Mark{3}ILLC &  \small\Mark{4}CIMeC\\
\small University of Groningen & \small Utrecht University & \small University of Amsterdam & \small University of Trento\tabularnewline
\end{tabular}\\
{\small\texttt{j.w.d.jumelet@rug.nl~~~e.g.bylinina@uu.nl}}\\{\small\texttt{w.h.zuidema@uva.nl~~~jakub.szymanik@unitn.it}}
}

\date{}

\begin{document}
\maketitle

\begin{abstract}
In English and other languages, multiple adjectives in noun phrases follow intricate ordering patterns. 
These patterns have been widely studied in linguistics and provide a useful test case for assessing how language models (LMs) acquire graded and context-sensitive word order preferences.
We ask to what extent adjective order preferences in LMs can be explained by distributional learning alone, and where models exhibit behaviour that goes beyond surface co-occurrence patterns.
We find that LM predictions are largely explained by training data frequencies: simple $n$-gram statistics account for much of their behaviour and closely mirror the preferences learned during training.
However, by analysing learning dynamics we reveal that models also generalize robustly to unseen adjective combinations, indicating that their behaviour cannot be reduced to memorization of observed orders alone.
Moreover, we show how LMs leverage word order cues from sentence context, demonstrating with feature attribution methods that contextual cues are an additional driver of adjective order in LM output. 

\end{abstract}
\section{Introduction}



\begin{figure}[t]
    \centering
    \includegraphics[width=\columnwidth]{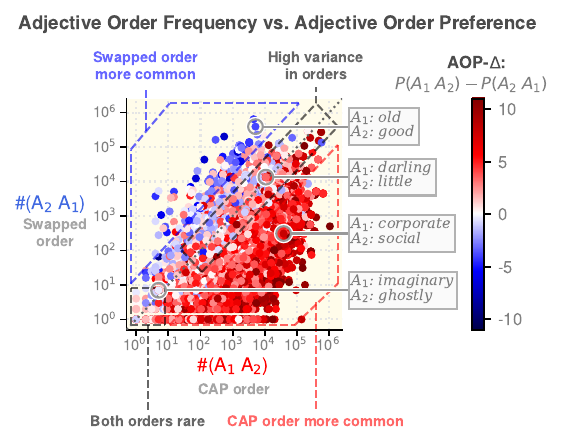}
    \caption{
    We connect the \textbf{adjective order preferences} (AOP-$\Delta$, \S\ref{sec:aop-description}) of language models (here Pythia-12b) to the \textbf{adjective order frequencies} of the corpus they have been trained on (The Pile).
    We highlight various regions of interest: adjective pairs for which both orders are rare and that require the LM to generalize from other adjective orders; pairs for which one particular order is far more common that can be resolved from frequency alone; and orders with high variance.
    }
    \label{fig:aop-map}
\end{figure}

Language models show impressive fluency in language generation and often score high on linguistic benchmarks, but the mechanisms underlying their linguistic proficiency remain hard to uncover. 
\newest{In particular, it remains unclear to what extent a model’s knowledge of word order reflects memorization of surface co-occurrence patterns, and to what extent it supports more generalizable behaviour.
What kinds of distributional and contextual information allow language models to go beyond observed word sequences when forming word order preferences?}

These questions have been posed and, in some cases, partially answered for various constructions and grammatical phenomena. 
The focus of such studies has mostly been on strict and binary grammatical constraints, where a violation results in an ungrammatical sequence. 
There has been less attention to how `softer' linguistic constraints are captured by LMs. 
Such constraints can give rise to slight and defeasible preferences. 
One example is adjective order in noun phrases where multiple adjectives modify one noun. 
Changing the order of adjectives does not necessarily lead to unacceptability, but can result in a slight decrease in naturalness. E.g.,  (`>' marks preference):

\begin{quote}\setlength\itemsep{0pt}
    \textit{large wooden box $>$ wooden large box}
\end{quote}

Moreover, these preferences are context-dependent: often, a context can be found where preferences switch -- for instance, when a contrasting property is introduced \citep{teodorescu2006adjective}:

\begin{enumerate}\setlength\itemsep{0pt}
    \item[ ] \textit{Take a \textsc{wooden} {large box}, not a plastic one.}
    \item[ ] \ $>$ \ \ \textit{... large wooden box ...}
\end{enumerate}

Such adjective order preferences ({\bf AOP}s) have been subject to \newest{extensive study} in theoretical linguistics, but have proved difficult to capture due to their graded and context-sensitive nature. 
A wide range of factors have been proposed as predictors of AOPs, ranging from superficial properties like adjective length to semantic and pragmatic properties to information-theoretic properties. 
Even when all those factors are combined, however, a significant portion of AOPs remains unaccounted for: the best-performing multi-factor models explain only about 71\% of human AOPs \citep{dyer-etal-2023-evaluating}. 

\newest{This gap highlights the complexity of adjective order preferences and motivates their use as a challenging test case for studying how language models form word order preferences.}
\newest{Rather than treating language models as explanatory accounts of adjective order, we analyse their behaviour in order to ask which aspects of their performance can be traced to distributional properties of the training data, and where additional effects such as generalization beyond observed data and sensitivity to sentence context arise.}
We conduct a series of experiments that explore AOP in LMs, its relation to properties of training data statistics, the extent to which LMs  generalize beyond memorization, and how sentence context shapes model preferences.
The contributions of our paper are the following:

\begin{itemize}[leftmargin=0.15in]
\item We provide the first systematic empirical analysis of adjective order preferences (AOPs) in large language models, showing that models trained on natural language data exhibit strong and consistent preferences for naturally occurring adjective orders. 
Using intermediate training checkpoints from the Pythia model suite, we characterize how these preferences emerge over the course of training and identify three distinct stages of acquisition (\S\ref{sec:aop_in_llms}).



\item To assess the role of training data statistics, we analyse $n$-gram distributions from the Pythia training corpus using Infini-gram \citep{liu2024infinigram}. 
We show that simple bigram statistics account for a large portion of AOP behaviour, and are strongly correlated with model predictions (\S\ref{sec:infinigram} and Figure~\ref{fig:aop-map}).

\item We then assess the extent to which model behaviour generalizes beyond memorization by evaluating AOP predictions for adjective combinations not seen \textit{during} training. Despite the strong influence of distributional statistics, models exhibit robust generalization to unseen adjective orders (\S\ref{sec:unseen_aop}).


\item Finally, we investigate the role of sentence context in shaping adjective order preferences. 
We show that contextual information systematically improves AOP predictions relative to neutral or random contexts, and use feature attribution methods to identify two sources of contextual influence: local collocational cues and earlier semantic cues that bias adjective ordering (\S\ref{sec:context}).

\end{itemize}

\noindent
Altogether, our experiments provide a behavioural characterization of how adjective order preferences in language models relate to training data statistics, generalization beyond memorization, and contextual information, and offer a general framework for studying word order in language models.

\section{Background}\label{sec:background}
\subsection{Adjective Order Theory}\label{sec:ling_background}
The mechanisms that drive adjective order have been a topic of research for decades (see \citealt{dyer-etal-2023-evaluating} for overview).
Theories have been proposed as far back as the 19\textsuperscript{th} century \citep{sweet1898new}, and refinements to the theory are being made to this day.
Early theories proposed lexical hierarchies that described adjective order based on abstract classes such as \textit{dimension} or \textit{color} \citep{Dixon1982, Sproat1991, Cinque_1996, scott2022}.
Adjectives were expected to be ordered based on the class they belong to: e.g. dimension goes in front of color, which gives rise to the order of a phrase like \textit{the big red house}.

While these hierarchies are powerful predictors of adjective order, they are not really explanatory: the question remains how such hierarchies are formed by the underlying properties of adjectives.
Furthermore, corpus studies have shown that many counterexamples against various versions of these hierarchies can be found \citep{truswell2009}, demonstrating the need for a more fine-grained analysis \citep{TrotzkeWittenberg}.

One line of work looks for semantic driving forces of AOPs, placing emphasis on adjectival properties like \textit{intersectivity} and \textit{subsectivity} 
\citep{truswell2009} or identifying conditions under which adjective order constraints can be lifted, such as \textit{focus} \citep{vendler1968adjectives,teodorescu2006adjective}.

Other approaches take a broader perspective, leveraging insights from quantitative linguistics, distributional semantics and information theory, and connecting this to psycholinguistic theories of human language processing.
Early analyses focus on properties such as \textit{length}---shorter adjectives tend to precede longer ones \citep{behaghel_von_1930}---and \textit{frequency}: more frequent adjectives tend to appear earlier \citep{MARTIN1969471, ney1983, wulff2003}.
This follows from a broader theory of language production, according to which more accessible and familiar material is produced first \citep{Bock1982TowardAC, FERREIRA2000296} and is comprehended faster \citep{ARNON201067}.
Collocational patterns of adjectives and nouns have been shown to be highly predictive of adjective order: adjectives most related to the noun appear closer to it \citep{sweet1898new, BYRNE197973, lapata-etal-1999-determinants}. 
This intuition has been operationalized through pointwise mutual information (\textit{PMI}) \citep{Hahn2018AnIE, futrell-etal-2020-determines}, which aligns with more general theories of word order arguing that distance between words that closely predict each other tends to be small \citep{GIBSON19981, Futrell2020DependencyLA}.

Predictors themselves can be complex, based on collocational or semantic factors or combinations thereof:  \textit{modification strength} of adjectives, for instance, is a predictor based on adjectival compositional potential \citep{Vecchi2013StudyingTR, Vecchi2017SpicyAA}; other predictors proposed in the literature include \textit{information gain} that an adjective provides for a noun \citep{dyer-etal-2021-predicting}, and \textit{subjectivity} of adjectives, with more subjective adjectives occurring first \citep{hill-2012-beauty,scontras2017,scontras2019,franke2019subjectivity}.

Although the majority of these approaches focus on English, various cross-lingual accounts have demonstrated their universal nature: adjective order tends to be driven by the same principles across languages \citep{wulff2015,leung-etal-2020-investigating,scontras2023adjective,dyer-etal-2023-evaluating}.

Finally, various papers highlight the need for a multi-factorial account of adjective order: it is unlikely that adjective order is driven by a single factor, and therefore it results from competing pressures \citep{wulff2003, TrotzkeWittenberg, dyer-etal-2023-evaluating}. Relatedly, a recent position paper \citep{levshina2023gradientorder} argues for a fundamentally gradient theory of word order that would be able to capture its hybrid and multi-faceted nature.


\newest{In our work, we study adjective order preferences in large language models as a means of analysing how such models handle graded and context-sensitive linguistic phenomena. 
Rather than evaluating language models as alternatives to existing linguistic theories of adjective order, we use them as a tool for probing which sources of information influence adjective ordering decisions.
In particular, while most theoretical accounts of adjective order focus on adjectives in isolation, language models naturally condition on rich sentential context. 
This makes them a useful testbed for investigating how contextual cues shape adjective order preferences. 
By analysing the contextual information that influences model predictions, we aim to identify patterns that may inform future empirical studies of adjective order and the role of context in human language use.}




\subsection{Word Order in Language Models}
With the increasing capabilities of language models, their linguistic abilities have been under much scrutiny in recent years.
A wide range of procedures have been introduced to uncover their understanding of grammaticality and notions of linguistic structure \citep{linzen2021baroni}.
Grammaticality is often assessed using the LM's probability predictions of minimal pairs of grammatical and ungrammatical sentences \citep{marvin-linzen-2018-targeted, warstadt-etal-2020-blimp-benchmark, gauthier-etal-2020-syntaxgym}; 
phenomena that have been examined this way include number agreement \citep{linzen-etal-2016-assessing, gulordava-etal-2018-colorless}, negative polarity items \citep{jumelet-hupkes-2018-language, bylinina-tikhonov-2022-transformers}, and filler-gap dependencies \citep{wilcox-etal-2018-rnn,suijkerbuijk-etal-2023-learnability}.
To our knowledge, no prior work has focused on adjective order in particular: the closest related phenomenon to this would be \citet{DBLP:journals/corr/abs-2403-19827}'s analysis of adjective-numeral constructions in LMs.

The importance of word order in LMs has been a topic of debate, with various works claiming that downstream performance is not affected by scrambled inputs \citep{malkin-etal-2021-studying, sinha-etal-2021-masked}, although it has been shown that LMs are able to retain a notion of word order through their positional embeddings \citep{abdou-etal-2022-word}.
It has been argued that LMs acquire a notion of word order that generalizes beyond $n$-gram co-occurrence statistics \citep{futrell-levy-2019-rnns, merrill2024evaluatingngramnoveltylanguage, DBLP:journals/corr/abs-2510-24963,DBLP:conf/iclr/0003AEAAZW25}, a claim that we in this paper assess for large-scale LMs in the context of adjective order.
Finally, numerous works have investigated the trade-off between memorization and generalization in LMs: it has been shown that larger LMs are able to memorize entire passages from the training data \citep{biderman2023, lesci2024causal, prashanth2024recitereconstructrecollectmemorization,DBLP:conf/iclr/WalLMSSZB25}, but generalization patterns for grammatical phenomena have also been shown to follow human-like generalization \citep{dankers-etal-2021-generalising, hupkes2023, alhama-etal-2023-linguistic}.

\section{Methods}\label{sec:methods}

\subsection{Measuring Word Order Preference}\label{sec:aop-description}
We measure a language model's AOP by comparing its log probability of the \textit{natural} order (the order in which a pair of adjectives appears in the sentence from our corpus) with a \textit{swapped} order.
In our experiments we measure AOP in two settings, either by considering the phrase in \textit{isolation}\footnote{We compute these isolated probabilities by prepending the noun phrase with `\textit{The}'.}, or by taking the \textit{context} of the phrase into account.
For a double adjective phrase $\textsc{a}_1 \textsc{a}_2 \textsc{n}$ that is extracted from a corpus with context \textsc{c}, we compute a model's preferred order as follows:
\begin{align}
    \textup{AOP-}\Delta(\textsc{a}_1 \textsc{a}_2 \textsc{n}) &= P(\textsc{a}_1 \textsc{a}_2 \textsc{n}) - P(\textsc{a}_2 \textsc{a}_1 \textsc{n}) \\
    \textup{AOP-}\Delta(\textsc{a}_1 \textsc{a}_2 \textsc{n}|\textsc{c}) &= P(\textsc{a}_1 \textsc{a}_2 \textsc{n}|\textsc{c}) - P(\textsc{a}_2 \textsc{a}_1 \textsc{n}|\textsc{c})
\end{align}
All probabilities are in log space.
The \aopDelta metric represents the \textit{magnitude} of a models' preference. 
In our experiments we also consider an accuracy metric based on \aopDelta, expressed as the number of items for which \aopDelta is positive:
\begin{align}
    \textup{AOP}\textit{-\%}(\mathcal{C}) =
    \frac{\left\{\phi\in\mathcal{C}\mid\textup{AOP-}\Delta(\phi) > 0\right\}}{|\mathcal{C}|}
\end{align}
Where $\mathcal{C}$ is an evaluation corpus of double adjective phrases, as described in the following section.
\citet{DBLP:journals/corr/abs-2403-19827} consider a similar metric for measuring properties of word order preferences in LMs, based on corrupted word orders.

\subsection{Evaluation Corpus}\label{sec:cap}
To evaluate adjective order preferences across a wide range of adjective pairs and contexts, \new{we make use of the double adjective corpus of \citet{jumelet2025misra}.
In our experiments we refer to this corpus as CAP: Corpus of Adjective Pairs.}
Earlier work has extracted adjective pairs from the Universal Dependencies treebank directly \citep{dyer-etal-2023-evaluating}, but we instead use CAP to collect model judgments across a wider range of adjective pairs.

\paragraph{Procedure}
The evaluation corpus is derived from the BabyLM 100M corpus \citep{charpentier-etal-2025-findings}, which consists of a mix of various English language sources including Wikipedia articles, books, and transcribed dialogue data.
From this corpus a pre-selection of sentences containing multiple adjectives is made by filtering all sentences containing at least two adjectives, based on a list of 11,000 adjectives released by \citet{futrell-etal-2020-determines}.
Next, dependency parse trees are generated for each sentence using \texttt{spaCy}'s RoBERTa-based parser.  
All sentences are then selected in which a \textsc{noun} contains two \textsc{amod} relations to an \textsc{adj} that is part of the adjective list.
{This procedure results in 9396 double adjective sentences, consisting of 2341 unique adjectives.}
A sample of sentences from the corpus is provided in Appendix~\ref{app:cap-examples}.

From these sentences we generate corrupted sentences in which the adjective order is swapped.
In case the first adjective is preceded by an indefinite article we ensure the article is changed accordingly to satisfy morpho-phonological constraints, 
for example:
\begin{enumerate}\setlength\itemsep{0pt}
    \item[$+$] \textit{Peter received \textbf{a} \textcolor{olive}{full athletic scholarship} [..]}
    \item[$-$] \textit{Peter received \textbf{an} \textcolor{red}{athletic full scholarship} [..]}
\end{enumerate}

We treat the corpus-attested order as a proxy for canonical preference, while acknowledging that adjective order preferences are graded and context-dependent

\begin{figure*}[t]
    \centering
    \includegraphics[width=5.23in]{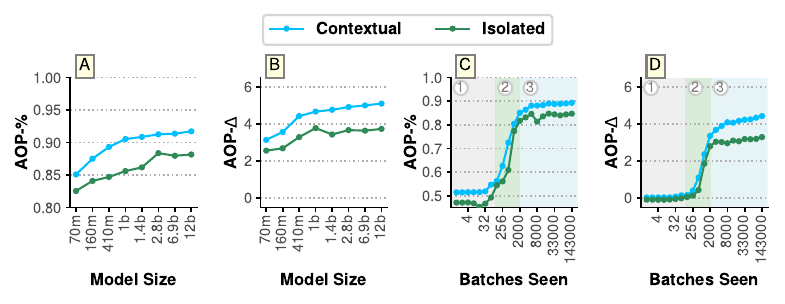}
    \caption{A--B: AOP-\% and AOP-$\Delta$ scores for Pythia models of increasing size. C--D: AOP-\% and AOP-$\Delta$ scores for Pythia-1.4b during training. We highlight the three learning phases: 1) initialization, 2) acquisition, and 3) consolidation.}
    \label{fig:model_size_training_dynamics}
\end{figure*}

\subsection{Models}
For our experiments we make use of the Pythia suite of language models \citep{biderman2023pythia}.
The models are released in increasing sizes (from 70M to 12B parameters) and all trained on The Pile corpus \citep{gao2020pile}.
Importantly, all intermediate checkpoints are available for these models in logarithmic intervals, which allows us to investigate the learning dynamics at a fine-grained level. 

\section{AOP in LMs}\label{sec:aop_in_llms}
In this first set of experiments we evaluate a series of pretrained LMs on CAP using the AOP-$\Delta$ and AOP-\% metrics of \S\ref{sec:aop-description}.

\paragraph{Model Size}
For the eight Pythia models of increasing size we plot the AOP\textit{-\%} performance in Figure~\ref{fig:model_size_training_dynamics}A, and the AOP\textit{-}$\Delta$ performance in Figure~\ref{fig:model_size_training_dynamics}B.
It can be seen that performance improves with model size: contextual AOP\textit{-\%} performance increases from 85.0\% for the 70m model to 91.7\% for the 12b model.
AOP-\% does not increase consistently with size, however, from the 1b model onward it reaches a plateau.
Interestingly, \aopDelta \textit{does} increase consistently with size: this shows that larger models become more \textit{certain} about their AOP judgment.

\paragraph{Learning Dynamics}
To investigate how AOP develops during training, we compute scores for the 1.4b model on all its intermediate checkpoints in Figure \ref{fig:model_size_training_dynamics}C and \ref{fig:model_size_training_dynamics}D.
From these plots we can identify three distinct learning phases: 1) an \textit{initialization} phase in which AOP judgments have not been formed yet, 2) an \textit{acquisition} phase in which AOP judgments are rapidly being formed, and 3) a \textit{consolidation} phase in which AOP judgments remain stable and only the role of context is being reinforced.
A similar tri-phase learning dynamic has been observed by \citet{van-der-wal-etal-2022-birth} and \citet{DBLP:journals/corr/abs-2309-07311}.
Note that the consolidation phase is reached relatively quickly: after around 2000 batches, which is \textasciitilde1.4\% of the total model training.
\newest{This suggests that sensitivity to adjective order preferences emerges early during model training.}
Furthermore, it highlights the importance of investigating learning dynamics at a logarithmic scale: with linear checkpoints we would not obtain this fine-grained level of precision.


\begin{figure}[t]
    \centering
    \includegraphics[width=2.94in]{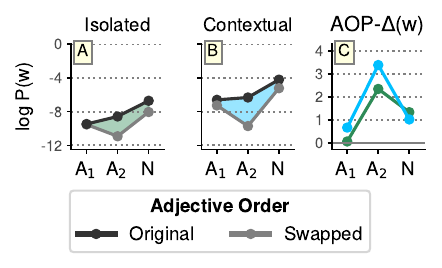}
    \caption{Average token probabilities for the original and swapped adjective orders on Pythia-12b, without and with sentence context (A--B), as well as the token-level differences (C) that correspond to the difference between the curves in (A) and (B).
    }
    \label{fig:token_probs}
\end{figure}

\paragraph{Localizing AOP}
In our experiments so far we have measured AOP based on the probabilities of the full adjective-noun triplet.
\new{Order preference} 
can arise at different locations of the phrase. 
The most natural location for this is 
the second adjective after having seen the first adjective, but context could also play a role for preferring which adjective comes at the first position, and a noun could be more probable after having seen two adjectives in their natural order.

In Figure \ref{fig:token_probs} we plot the average token probabilities on CAP for Pythia-12b, in isolation and in context.
We observe that, as expected, the largest probability difference occurs at the position of the second adjective: the model is highly surprised by encountering the second swapped adjective after having seen the first.
The noun probability is strongly affected by swapping adjective order as well: a noun is more likely to follow adjectives in natural order.
Furthermore, we can see that context plays a role for the probability of the first adjective already: whereas in the isolated case both adjectives in first position receive equal probability, order preference already manifests here when the model has access to context.

\paragraph{Conclusion} 
The Pythia LMs acquire a strong sense of adjective order that is learned early in training at around 1.5\% of the total amount of training.
We showed that sentence context has a positive effect on AOP, bumping up AOP-\% from 88\% to around 91\%.
By localizing the AOP scores on the token-level we show how these order preferences manifest at different points in a phrase.
In the next experiments we will investigate these results in more detail, connecting LM AOPs to training data statistics.

\section{The Role of Training Data Statistics in AOP}\label{sec:wimbd}
Having established that language models achieve high accuracy on adjective order prediction, we now examine the extent to which this behaviour can be explained by distributional properties of the training data.
Specifically, we ask to what extent simple co-occurrence statistics explain model predictions, and where systematic deviations from such statistics arise.

\newest{To ensure that our subsequent analyses are not specific to a single model family, we first examine adjective order behaviour across a range of contemporary large language models.}
Table ~\ref{tab:cap_accuracy} shows the performance of four contemporary, mid-size, open-source LLMs: Llama3-8b, OLMo2-7b, OLMo3-7b and Qwen3-8b \citep{grattafiori2024llama3herdmodels, yang2025qwen3technicalreport, olmo2025olmo3, olmo20252olmo2furious}. 
These models perform similarly to Pythia (slightly worse, at 89 to 91\%) with high \aopDelta correlation, supporting the use of the Pythia models as a representative case in our subsequent analyses.

\newest{As a next step, we assess whether LM adjective order behaviour can be accounted for by cognitive predictors proposed in prior work (\S\ref{sec:ling_background}).}
Compared to such predictors, \newest{language models exhibit adjective order behaviour that is not well explained by any single factor.}
In Table~\ref{tab:cap_accuracy} we show the accuracy of two predictors: PMI between adjective and noun (62.4\%), and adjective subjectivity (69.3\%).
As show in Appendix~\ref{app:cognitive-predictors}, these predictors show limited overlap with the \aopDelta scores, indicating that LM behaviour cannot be straightforwardly reduced to the effects captured by these individual predictors.

\begin{table}[t]
\centering
\footnotesize
\arrayrulecolor[rgb]{0.753,0.753,0.753}
\ADLnullwidehline
\addtolength{\tabcolsep}{-0.3em}
\begin{tabular}{lccc} 
Metric & AOP-\% & $\rho(\textup{AOP-}\Delta(\bullet))$ & $\rho(\textup{AOP-}\Delta(\bullet|c))$  \\[2pt]\arrayrulecolor{black}\hline\addlinespace[3pt]
AOP-$\Delta(\bullet)$                                                                    & 88.1     & ---                                     & 0.67                                    \\
AOP-$\Delta(\bullet|c)$                                                         & \textbf{91.7}     & 0.67                                 & ---                                      \\[3pt]\arrayrulecolor[rgb]{0.753,0.753,0.753}\hdashline[1pt/1pt]\addlinespace[3pt]
PMI(\textsc{a};\textsc{n})                           & 62.4     & 0.27                                   & 0.20                                    \\
Subj(\textsc{a})                        & \textbf{69.3}     & 0.07                                   & 0.08                                    \\[3pt]\arrayrulecolor[rgb]{0.753,0.753,0.753}\hdashline[1pt/1pt]\addlinespace[3pt]
$\%(\textsc{a}_1)$                                    & 55.2     & 0.10                                   & 0.13                                    \\
$\%(\textsc{c}$$\cdot$$\textsc{a}_1)$                                & 61.7     & 0.11                                   & 0.22                                    \\
$\%(\textsc{a}_1\textsc{a}_2)$                                & \textbf{86.8}     & 0.56                                   & 0.55                                    \\
$\%(\textsc{a}_2\textsc{n})$                                & 60.6     & 0.16                                   & 0.09                                    \\
$\%(\textsc{a}_1\textsc{a}_2\textsc{n})$                                & 76.7     & 0.49                                   & 0.48                                    \\[3pt]\arrayrulecolor[rgb]{0.753,0.753,0.753}\hdashline[1pt/1pt]\addlinespace[3pt]
Llama3-8b & 89.0 & 0.63 & 0.76\\
OLMo2-7b & 90.6 & 0.67 & 0.84 \\
OLMo3-7b & 89.5 & 0.78 & 0.82 \\
Qwen3-8b & 89.2 & 0.74 & 0.80 \\
\arrayrulecolor{black}\bottomrule
\end{tabular}
    \caption{Predictive accuracy of adjective order on CAP for various metrics: \aopDelta for Pythia-12b 
    and relative $n$-gram counts from The Pile (\S\ref{sec:wimbd}).
    The highest accuracy for each group is \textbf{bolded}.
    For each metric we provide the Spearman correlations with respect to the \aopDelta scores of Pythia-12b.
    AOP scores for the four LLMs denote contextual AOP.
    }
    \label{tab:cap_accuracy}
\end{table}

\subsection{Infini-gram}\label{sec:infinigram}
In order to better understand the factors underlying LM adjective order behaviour, we examine how double adjective constructions are distributed in the training data.
The way that LMs are trained nowadays makes this challenging: many models are released without open access to their training data, and the scale of training corpora that \textit{are} available poses a challenge for targeted analysis.

To overcome this, we make use of the Infini-gram API \citep{liu2024infinigram}, which provides a fully indexed search engine over various large-scale corpora such as The Pile, on which the Pythia models have been trained.
Infini-gram makes it possible to retrieve the exact $n$-gram statistics that LMs have been trained on, and as such provides a detailed insight into the collocational factors that play a direct role in shaping AOP.

We retrieve the uni-, bi- and trigram counts for all the adjective-noun triplets in CAP for The Pile, as well as for the swapped orders.
To provide an analogue to \aopDelta, we express the $n$-gram counts as log count differences.
We express these relative $n$-gram counts as 
\[\%(\textsc{a}_1\textsc{a}_2\textsc{n})=\log \#(\textsc{a}_1\textsc{a}_2\textsc{n})-\log \#(\textsc{a}_2\textsc{a}_1\textsc{n})\]
with similar formulations for adjective bigrams (\%($\textsc{a}_1\textsc{a}_2$)), adjective-noun bigrams (\%($\textsc{a}_2\textsc{n}$)), bigrams of the last context token and adjective (\%($\textsc{c}$$\cdot$$\textsc{a}_1$)) and unigrams (\%($\textsc{a}_1$)).

\paragraph{Predictive Accuracy}
We present the predictive accuracy of these relative $n$-gram occurrences in Table~\ref{tab:cap_accuracy}.
\%(\textsc{a}$_1$) predicts adjective order with \new{55.2\%}, which provides weak evidence that more frequent adjectives tend to occur first \citep{MARTIN1969471, TrotzkeWittenberg}.
\%(\textsc{a}$_1$\textsc{a}$_2$) is the strongest predictor of adjective order at \new{86.8}\%.
This shows that a large majority of the items in CAP can be resolved based on bigram statistics alone, which is leveraged by the LMs during training.
However, $n$-gram collocations alone do not fully explain LM performance: a large amount of trigrams are not part of the Pile and performance for $\%(\textsc{a}_1\textsc{a}_2\textsc{n})$ drops to \new{76.7}\%.

\begin{figure}
    \centering
    \includegraphics[width=0.99\columnwidth]{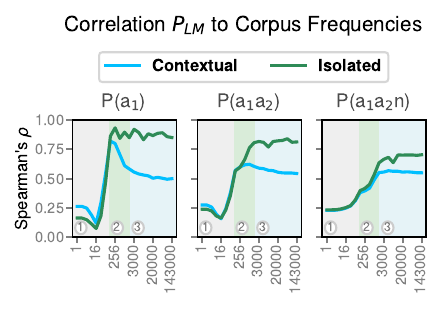}
    \caption{Correlations during training of LM probabilities for single adjectives, adjective pairs, and adjective-noun triplets with respect to their frequency in The Pile.
    }
    \label{fig:frequency-corr}
\end{figure}
\paragraph{Correlations}
To investigate how $n$-gram statistics 
\new{relate to} the formation of \aopDelta during training, we compute the Spearman correlation between various log-transformed $n$-gram counts and a LM's \aopDelta on intermediate checkpoints.
We plot this in Figure~\ref{fig:frequency-corr}, for contextual and isolated \aopDelta.
The trends we observe here align closely with the three learning phases that we outlined in \S\ref{sec:aop_in_llms}.
The initialization phase (1) is marked by a rapid acquisition of unigram and bigram statistics, that remain unaffected by context at this point.
Then, at the start of the acquisition phase (2), in which AOP performance rapidly improves, the model starts to incorporate linguistic context in its uni- and bigram predictions.
Finally, in the consolidation phase (3), trigram predictions start being affected by context as well, whereas unigram and bigram predictions remain stable.
The fully trained model strongly reflects these correlations to data statistics, which explains the strong relation we observed in Figure~\ref{fig:aop-map} between bigram frequencies and \aopDelta.
\new{This successive alignment from unigram to higher $n$-gram statistics has also been observed in prior work \citep{10.1162/tacl_a_00444, DBLP:journals/corr/abs-2510-24963}.}

\subsection{Generalizing to Unseen Adjective Orders}\label{sec:unseen_aop}
Having access to the $n$-gram distribution a model was trained on allows us to investigate its behaviour on adjective orders not previously observed during training.
We ask whether models exhibit systematic generalization to unseen adjective combinations, beyond direct memorization of observed co-occurrence patterns.
However, due to the enormous size of The Pile (300B tokens) it turns out that number of unseen adjective pairs in CAP is too small to draw significant conclusions from (only 39 out 9396 pairs, 0.4\%).
This leaves us with two options: either we collect an additional sample of rare adjective pairs that are not part of the Pile, or we investigate the model at intermediate checkpoints where a greater amount of adjective pairs has not been encountered yet.
An issue with collecting highly rare pairs is that their canonical order will be much harder to determine, for which we would require extensive human judgments on the most natural order.
We therefore go for the second option, and set up a procedure to measure the data statistics during training.

\begin{figure}
    \centering
    \includegraphics{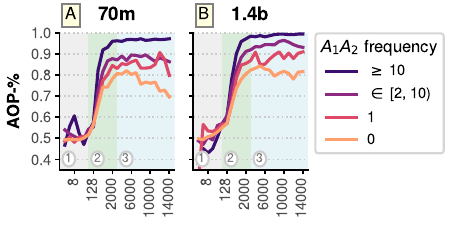}
    \caption{
        The contextual AOP-\% performance for Pythia-70m and 1.4b across training, split out for items that have been seen 0, 1, 2 to 10, and more than 10 times at each specific checkpoint.
        We provide the size of these splits across training in Appendix~\ref{app:split-sizes}.
    }
    \label{fig:ngram-curve}
\end{figure}
\paragraph{Intermediate $n$-gram Counts}
Since Infini-gram only provides $n$-gram statistics for The Pile as a whole, we need to collect the intermediate statistics ourselves.
Using the Pythia batch viewer \citep{biderman2023pythia} this becomes possible: our implementation is able to process the first 10\% of The Pile (up to batch 14,000; 29.4B tokens) under one hour on a consumer-grade laptop. 
We collect the bi-gram counts of all adjective pairs in CAP on the \textit{batch} level, allowing us to determine exactly when and how often a specific adjective pair has been observed during training. 

\paragraph{Unseen Adjective Pairs}
Based on these intermediate $n$-gram counts, we compute the AOP-\% scores for various splits based on the number of times an adjective pair has been encountered.
We focus on the cases where the swapped order has \textit{never} been encountered: this allows us to isolate the moment a model encounters an adjective combination for the first time, and measure its `zero-shot' performance on these cases. 
For each model checkpoint we collect the $n$-gram counts at that particular point in training, and split the CAP items based on the number of times an \textsc{a}$_1$\textsc{a}$_2$ bi-gram has been seen.

We present the results for Pythia-70m and 1.4b in Figure~\ref{fig:ngram-curve}.
The performance on unseen pairs is remarkably high, reaching around 85\% at the highest point. 
We provide a more detailed item-level analysis for these items in Appendix~\ref{app:one-shot-aop}: while it is clear on the split level that a single occurrence improves AOP-\%, we do not observe such a clear effect on the item level.
Interestingly, as the number of unseen adjective pairs decreases, the performance on unseen pairs decreases as well.
\newest{As training progresses and the pool of unseen adjective pairs becomes smaller, performance on these unseen items decreases.
Since the set of unseen pairs at later checkpoints is a strict subset of earlier unseen sets, this trend likely reflects increasing item difficulty rather than a qualitative change in model generalization.}

\paragraph{Conclusions}
The question we set out to address in this set of experiments is the extent to which adjective order behaviour in language models can be explained by direct co-occurrence statistics.
We approached this question by analysing the overall $n$-gram distribution of the Pile and its evolution over the course of training.
On the one hand, AOP is strongly connected to training data statistics: we observe a strong correlation between AOP-$\Delta$ and count-based AOP scores, as visualized in Figure~\ref{fig:aop-map}.
On the other hand, the relatively strong performance on adjective pairs not previously observed during training indicates that \newest{models generalize adjective order behaviour beyond direct memorization of seen co-occurrences.}

\begin{figure}
    \centering
    \includegraphics[]{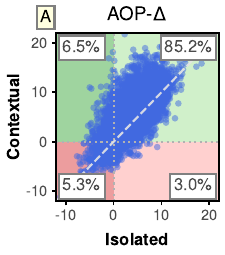}\hfill
    \includegraphics[]{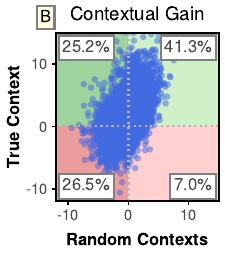}
    \caption{
    A: Item-level \aopDelta scores with and without sentence-context for Pythia-12b. 
    We provide a sample of sentences from each quadrant in Appendix~\ref{app:cap-examples}.
    B: The contextual gain ($c$-$\Delta$, Eq.~\ref{eq:c-delta}) of an item's true context compared to an expectation over random contexts (Eq.~\ref{eq:random-aop}).
    }
    \label{fig:role_of_context}
\end{figure}
\section{The Role of Context}\label{sec:context}
In \S\ref{sec:aop_in_llms} we established that LMs have a thorough grasp of adjective order that manifests itself early during training.
We showed that having access to sentence context boosts performance, and we now investigate the role of context in more detail.


\subsection{Quantifying Context Impact}
\paragraph{Context improves predictions a lot}
To understand in what ways AOP is impacted by context, we investigate \aopDelta scores on an item-level with and without context.
In Figure~\ref{fig:role_of_context}A, we plot the \aopDelta scores in isolation against the contextual scores, and mark the four quadrants of items that result in a positive or negative score after adding context.
In the majority of cases (85.2\%), \aopDelta is positive both with and without context: the model is able to determine adjective order based on the adjective-noun triplet alone.
In 6.5\% of the cases, however, adding context flips a negative \aopDelta score to a positive one.
Adding context has a boosting effect in general: in 67.6\% of the cases, \aopDelta is increased by adding context.
We provide an overview of how these ratios are affected by model size in Appendix~\ref{sec:app_context_ratios}: the number of cases where context improves AOP-\% is actually \textit{larger} for small models, but so is the number of cases where context worsens it.



\begin{figure}
    \centering
    \includegraphics{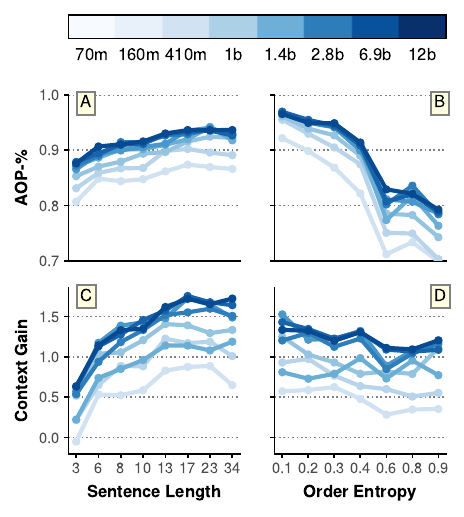}
    \caption{Contextual AOP-\% and Context Gain against Sentence Length and (Adjective) Order Entropy for all Pythia models.
    Scores are binned based on quantiles: $x$-tick labels indicate the lower bound of each bin (e.g. $[3,6)$ for the first tick). 
    }
    \label{fig:len_entropy}
\end{figure}

\paragraph{Random Contexts}
The fact that context improves \aopDelta raises the question what information in the context is responsible for this improvement.
Does context contain relevant semantic information that plays a meaningful role for determining adjective order, or is any linguistic signal sufficient for this?
To test this, we compute an expectation over \aopDelta scores with \textit{random contexts}, in which an adjective pair is preceded by a randomly sampled context from another adjective pair in CAP:
\begin{equation}\label{eq:random-aop}
\textup{AOP-}\Delta(\textsc{a}_1\textsc{a}_2\textsc{n}\mid\mathcal{C}) =  \mathbb{E}_{c\sim\mathcal{C}}\left[\textup{AOP-}\Delta(\textsc{a}_1\textsc{a}_2\textsc{n}\mid c)\right]    
\end{equation}
We compute the expectation over a random sample of 500 contexts, taken from CAP ($\mathcal{C}$).
To express the \textit{relative} impact of adding context, we regress out a model's isolated \aopDelta score by subtracting it from the contextual scores, which we call the \new{\textbf{contextual gain}} ($c\textup{-}\Delta$):
\begin{equation}\label{eq:c-delta}
    c\textup{-}\Delta(\bullet|c) = \textup{AOP-}\Delta(\bullet|c) - \textup{AOP-}\Delta(\bullet)
\end{equation}
By computing contextual gain for both the original context and for the expectation over random contexts we can quantify how much more important the original context is for \aopDelta.

We plot the results for this experiment in Figure~\ref{fig:role_of_context}B.
The true context improves isolated \aopDelta in 65.1\% of the cases, whereas random contexts lead to an improvement for only 39.6\% (for 33.2\% only the true context leads to an improvement whereas random contexts don't). 


\begin{figure*}[t!]
    \centering
    \includegraphics[]{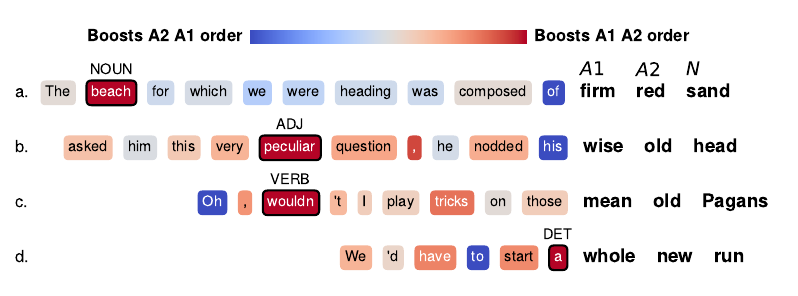}
    \caption{Four sentences annotated with the Integrated Gradient attribution scores, that each denote a specific different way in which part-of-speech plays a role in contextual adjective order.}
    \label{fig:attribution_examples}
\end{figure*}
\paragraph{Context Properties}
In Figure \ref{fig:len_entropy} we plot the impact of context length against AOP-\% and Context Gain.
Longer contexts lead to better performance as a whole (\ref{fig:len_entropy}A), and lead to larger differences between the contextual and isolated AOP scores (\ref{fig:len_entropy}C).
This effect gradually increases with model size: longer contexts have the greatest effect for the largest model.

One way that contextual cues may help a model is to disambiguate adjective orders with high entropy.
We compute adjective order entropy based on the $\textsc{a}_1\textsc{a}_2$ bigram counts from Infini-gram (\S\ref{sec:infinigram}).
In Figure \ref{fig:len_entropy}B we demonstrate that model performance drops significantly for adjective pairs with high entropy: these are the hardest cases to get right.
Surprisingly, however, the {context gain} does \textit{not} increase for high entropy items (\ref{fig:len_entropy}D), indicating that the model can not leverage contextual cues for disambiguation.

\subsection{Uncovering Context Cues}
Thus far we have only examined the role of context on a macro-level.
We now present an approach to identify specific words in the context that are of importance to the model's predictions.
For this we make use of \textbf{feature attribution methods}, which express the relative importance of input features to a prediction.

\paragraph{Procedure}
Our procedure for identifying salient context cues is inspired by \citet{DBLP:conf/iclr/SartiCNB24}, who use feature attribution techniques to identify contextual cues in machine translation.
Key to this approach is the usage of \textit{contrastive explanations} \citep{yin-neubig-2022-interpreting}, where the contribution of a context token is expressed with respect to the \textit{difference} of two possible outputs (in our case the two adjective orders).
\begin{enumerate}
     \item Using the context gain scores of Eq.~\ref{eq:c-delta}, we create a subset of context-dependent constructions by taking the top 20\% of items in CAP. These are the items for which the sentence context had the greatest impact on the model's adjective order preference.
     \item We compute the contribution of the tokens in the context using Integrated Gradients \citep{ig2017} with respect to the AOP-$\Delta$ scores.\footnote{\new{Scores are obtained using \texttt{inseq} \citep{sarti-etal-2023-inseq}, with 0-valued baselines.} We \textit{sum} over the embedding dimension to aggregate subword-level attributions, and sum subword attributions to obtain token-level attributions. Since Integrated Gradients is \textit{linear} we can obtain contrastive explanations by subtracting the attribution scores of the swapped phrase from the scores of the original phrase.} Positive contribution scores indicate that that token boosts the original order, whereas a negative score indicates a contribution to the opposite order.\footnote{\new{Feature attribution methods have been under extensive scrutiny in recent years: the complexity of large-scale models makes it challenging to capture their behaviour in a singular importance score, which will lead to disagreement between different methods \citep{DBLP:conf/hhai/NeelySBL22}. Integrated Gradients have nonetheless been shown to be reliable estimators of feature importance \citep{10543008, DBLP:conf/iclr/SartiCNB24}, and therefore serve as a useful baseline for our experimental setup. \newest{We acknowledge though that different attribution methods may yield different insights.}}}
     \item We filter for cases where the top contributing token had a score of at least 1, indicating that there is a salient cue present in the context.
     \item In order to map our results to a `global' interpretation we POS tag the contexts using SpaCy's \texttt{en\_core\_web\_lg} tagger \citep{honnibal2020spacy}. This allows us to express the model behaviour in terms of POS classes, rather than on a lexical level.
 \end{enumerate}

\begin{figure}[t!]
    \centering
    \includegraphics[width=2.6in]{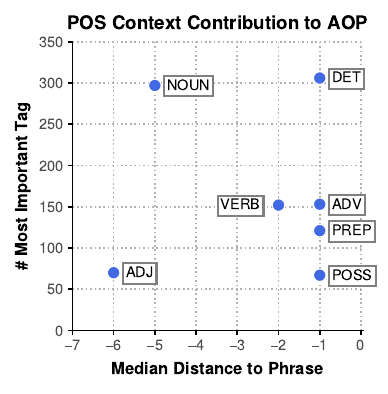}
    \caption{
    Distribution of the most important token in the context, grouped by POS tag.
    Each point indicates the number of times that POS tag was the most important token, and the median (relative) distance to the adjective phrase.
    POS tags with a frequency below 30 are filtered out.
    }
    \label{fig:aop_attribution_pos}
\end{figure}
\paragraph{Results}
We perform our procedure on the Pythia-1.4b model.
In Figure~\ref{fig:attribution_examples}, we show four annotated sentences that are sampled randomly from the corpus.
Each sentence exemplifies the way different part-of-speech classes can play a role in influencing contextual AOP.
It can be seen that a context may contain `competing' cues that prefer the natural and swapped adjective order, highlighting that context can impact model behaviour in different ways.

To aggregate this result, we measure the amount of times a POS tag was the most important token in the context, and its distance from the adjective phrase.
We visualize this in Figure~\ref{fig:aop_attribution_pos}.
From this plot we conclude that there are two types of contextual cues that play a role:
\begin{enumerate}
    \item \textbf{Collocational cues}: Determiners, adverbs, and prepositions directly preceding the adjective can boost a specific order due to collocational tendencies 
    playing a role. \new{This is another way for $n$-gram statistics to play a role in AOP -- for instance, adjectives that tend to occur in the left periphery of the noun phrase would co-occur with determiners more often.}
    The Infini-gram scores for the context-adjective bigrams show that the last context token \textsc{c} is a strong driver of adjective order here: $\textsc{c}$$\cdot$$\textsc{a}_1$ is more frequent than $\textsc{c}$$\cdot$$\textsc{a}_2$ in 84.0\%---for CAP as a whole this is only 61.7\%---demonstrating that bigram statistics act as an important cue here for the LM to prefer a prediction of $\textsc{a}_1$ directly after  \textsc{c}.
    \item \textbf{Semantic cues}: 
    \new{Non-immediate context can play a role in AOP under a different mechanism. Our results highlight the role of adjectives and nouns that occur quite early in the sentence.} 
    In the case of adjectives this is strongly driven by semantic similarity: top-scoring adjectives in the context have on average a cosine similarity of 0.15 with the first adjective in the phrase, and only 0.07 with the second adjective.\footnote{Cosine similarity is computed using the input embeddings of the LM, summing subword embeddings where necessary.}
\end{enumerate}

\subsection{Conclusions}
The results in this section show that adjective order behaviour in LMs is \textbf{context-sensitive}.
Building on the aggregate effects reported earlier (Figure~\ref{fig:token_probs}C), the attribution analysis identifies two classes of contextual cues that shape model preferences: local collocational cues in the immediate vicinity of the adjective phrase, and earlier semantic cues that bias adjective ordering based on semantic similarity.

\newest{Importantly, these findings do not constitute an explanatory account of adjective order in human language.
Rather, they illustrate how language models can serve as a tool for probing which aspects of sentence context are informative for adjective order preferences.
In doing so, this analysis highlights contextual factors that are largely absent from existing theories of adjective order, and may serve as a starting point for future empirical studies on the role of context in human adjective order preferences.}

\section{Discussion}\label{sec:discussion}
\paragraph{Generalization and Memorization in AOP}
In this paper, we investigated the extent to which language models generalize adjective order preferences beyond direct memorization of training data statistics.
Our results suggest that while models exhibit some degree of generalization, this generalization is limited and coexists with a strong reliance on distributional information.
Here, we want to highlight an important question that is rarely brought up in linguistic generalization discussions in NLP: How general and abstract is human linguistic behaviour? 

The linguistic literature suggests that human language production, including word order preferences, is guided by a mix of memorization and abstract principles \citep{c0f64ae7-9422-34a7-b044-fa63660e7879}. 
Speakers tend to recycle more routinised and therefore more accessible combinations \citep{macdonald2013language}, guided by sequence frequencies in linguistic input that, in turn, form sequential associations that lead to production automatizing \citep{bybee2010language, diessel2019grammar}. 
This is compatible with a picture of language acquisition where knowledge of grammar is formed based on encountered exemplars of usage, which are stored in memory \citep{bybee2010language} and serve as basis of generalization \citep{goldberg2006constructions}, but can also serve as patterns for routinised language production. 
So, human linguistic knowledge and language production is not totally abstract and is not based exclusively on generalization. 
This has been discussed across linguistic phenomena and has been suggested for word order \citep{macdonald2013language, levshina2023gradientorder}, although not for adjective order specifically. 

When investigating generalization in LMs it is therefore important to make a distinction between the degree to which they create generalizations for a particular phenomenon, and how similar this generalization is to human behaviour. 
In principle, one could even find cases of LM \textit{overgeneralization}. 
Importantly, in order to answer how human-like a model's generalization behaviour is, we would need data on the trade-off in memorization and generalization in human linguistic behaviour: an empirical question for which detailed human data on adjective order preferences is currently lacking.

\paragraph{Corpus Interventions}
Our analysis of AOP generalization is constrained by the fixed training corpus the models are trained on.
More detailed insights into this question can be obtained by an intervention-based approach, in which we filter out specific constructions from the training data \citep{jumelet-etal-2021-language,DBLP:journals/corr/abs-2403-19827, patil2024filtered,muller-eberstein-etal-2025-decaf}.
An interesting experiment in this direction would be to remove \textit{all} sentences containing double adjectives from the training corpus, and measure whether the remaining distribution of adjectives contains a sufficient signal to acquire adjective order constraints.

\paragraph{Cross-lingual Extensions}
Our analysis is limited to adjective order preferences in English, which exhibits ordering patterns that differ from those found in other languages.
Cross-linguistic research has documented both variation and regularities in adjective ordering, motivating general theories that go beyond any single language \citep{cinque2010syntax, culbertson2020}.
We leave a systematic cross-linguistic investigation of adjective order preferences in language models to future work.

\paragraph{LMs and Linguistic Theory}
Despite the impressive performance of LMs in recent years, their impact on theoretical linguistics has remained minimal \citep{baroni2022}.
There are various reasons for this.
On the one hand, the cognitive plausibility of current LMs remains questionable, both in terms of the scale at which they are trained \citep{huebner-etal-2021-babyberta, WILCOX2025104650} and their inductive biases that depend strongly on model architecture and are not reflective of human language processing \citep{wilson-frank-2023-inductive, oh-etal-2024-frequency, padovani-etal-2025-child}.
On the other hand, their black box nature makes it highly challenging to extract linguistic patterns that can be tested against existing linguistic theories.

\newest{
In this paper, we treat language models not as explanatory theories of adjective order, but as empirical tools for analysing how soft word order constraints are acquired by LMs.
Although language models substantially outperform individual cognitive predictors proposed in prior work, predictive accuracy alone does not constitute an explanation of adjective order preferences in human language.
Instead, we view their value as lying in the combination of behavioural analysis with \textbf{interpretability} methods, which makes it possible to probe what sources of information influence model decisions \citep{Futrell_2025}.
In particular, the attribution-based analysis in \S~\ref{sec:context} illustrates how such methods can be used to identify contextual cues that systematically modulate adjective order preferences---factors that are largely absent from existing accounts and that may serve as targets for future empirical investigation in human language.
}


\section*{Acknowledgements}
We gratefully thank the anonymous reviewers for their feedback on an earlier version of the paper, which greatly helped in shaping the story.
We thank Kanishka Misra, Kyle Mahowald and Jennifer Culbertson for the inspiration that our adjacent collaboration has provided for finalizing this paper.
Jaap Jumelet is supported by NWO grant \texttt{VI.Vidi.221C.009}. 


\bibliography{custom}

@book{goldberg2006constructions,
  title={Constructions at Work: The Nature of Generalization in Language},
  author={Goldberg, A.E.},
  isbn={9780199268511},
  lccn={2006296206},
  series={Oxford linguistics},
  year={2006},
  publisher={Oxford University Press}
}

@book{bybee2010language,
  title={Language, usage and cognition},
  author={Bybee, Joan},
  year={2010},
  publisher={Cambridge University Press}
}

@inproceedings{charpentier-etal-2025-findings,
    title = "Findings of the Third {B}aby{LM} Challenge: Accelerating Language Modeling Research with Cognitively Plausible Data",
    author = "Charpentier, Lucas  and
      Choshen, Leshem  and
      Cotterell, Ryan  and
      Gul, Mustafa Omer  and
      Hu, Michael Y.  and
      Liu, Jing  and
      Jumelet, Jaap  and
      Linzen, Tal  and
      Mueller, Aaron  and
      Ross, Candance  and
      Shah, Raj Sanjay  and
      Warstadt, Alex  and
      Wilcox, Ethan Gotlieb  and
      Williams, Adina",
    editor = "Charpentier, Lucas  and
      Choshen, Leshem  and
      Cotterell, Ryan  and
      Gul, Mustafa Omer  and
      Hu, Michael Y.  and
      Liu, Jing  and
      Jumelet, Jaap  and
      Linzen, Tal  and
      Mueller, Aaron  and
      Ross, Candace  and
      Shah, Raj Sanjay  and
      Warstadt, Alex  and
      Wilcox, Ethan Gotlieb  and
      Williams, Adina",
    booktitle = "Proceedings of the First BabyLM Workshop",
    month = nov,
    year = "2025",
    address = "Suzhou, China",
    publisher = "Association for Computational Linguistics",
    url = "https://aclanthology.org/2025.babylm-main.28/",
    doi = "10.18653/v1/2025.babylm-main.28",
    pages = "399--420",
    ISBN = "TODO"
}

@inproceedings{padovani-etal-2025-child,
    title = "Child-Directed Language Does Not Consistently Boost Syntax Learning in Language Models",
    author = "Padovani, Francesca  and
      Jumelet, Jaap  and
      Matusevych, Yevgen  and
      Bisazza, Arianna",
    editor = "Christodoulopoulos, Christos  and
      Chakraborty, Tanmoy  and
      Rose, Carolyn  and
      Peng, Violet",
    booktitle = "Proceedings of the 2025 Conference on Empirical Methods in Natural Language Processing",
    month = nov,
    year = "2025",
    address = "Suzhou, China",
    publisher = "Association for Computational Linguistics",
    url = "https://aclanthology.org/2025.emnlp-main.999/",
    doi = "10.18653/v1/2025.emnlp-main.999",
    pages = "19735--19756",
    ISBN = "979-8-89176-332-6"
}

@article{WILCOX2025104650,
title = {Bigger is not always better: The importance of human-scale language modeling for psycholinguistics},
journal = {Journal of Memory and Language},
volume = {144},
pages = {104650},
year = {2025},
issn = {0749-596X},
doi = {https://doi.org/10.1016/j.jml.2025.104650},
url = {https://www.sciencedirect.com/science/article/pii/S0749596X25000439},
author = {Ethan Gotlieb Wilcox and Michael Y. Hu and Aaron Mueller and Alex Warstadt and Leshem Choshen and Chengxu Zhuang and Adina Williams and Ryan Cotterell and Tal Linzen}
}

@inproceedings{DBLP:conf/iclr/SartiCNB24,
  author       = {Gabriele Sarti and
                  Grzegorz Chrupala and
                  Malvina Nissim and
                  Arianna Bisazza},
  title        = {Quantifying the Plausibility of Context Reliance in Neural Machine
                  Translation},
  booktitle    = {The Twelfth International Conference on Learning Representations,
                  {ICLR} 2024, Vienna, Austria, May 7-11, 2024},
  publisher    = {OpenReview.net},
  year         = {2024},
  url          = {https://openreview.net/forum?id=XTHfNGI3zT},
  bibsource    = {dblp computer science bibliography, https://dblp.org}
}

@inproceedings{ig2017,
author = {Sundararajan, Mukund and Taly, Ankur and Yan, Qiqi},
title = {Axiomatic attribution for deep networks},
year = {2017},
publisher = {JMLR.org},
booktitle = {Proceedings of the 34th International Conference on Machine Learning - Volume 70},
pages = {3319–3328},
numpages = {10},
location = {Sydney, NSW, Australia},
series = {ICML'17}
}

@inproceedings{sarti-etal-2023-inseq,
    title = "Inseq: An Interpretability Toolkit for Sequence Generation Models",
    author = "Sarti, Gabriele  and
      Feldhus, Nils  and
      Sickert, Ludwig  and
      van der Wal, Oskar",
    editor = "Bollegala, Danushka  and
      Huang, Ruihong  and
      Ritter, Alan",
    booktitle = "Proceedings of the 61st Annual Meeting of the Association for Computational Linguistics (Volume 3: System Demonstrations)",
    month = jul,
    year = "2023",
    address = "Toronto, Canada",
    publisher = "Association for Computational Linguistics",
    url = "https://aclanthology.org/2023.acl-demo.40/",
    doi = "10.18653/v1/2023.acl-demo.40",
    pages = "421--435"
}

@inproceedings{yin-neubig-2022-interpreting,
    title = "Interpreting Language Models with Contrastive Explanations",
    author = "Yin, Kayo  and
      Neubig, Graham",
    editor = "Goldberg, Yoav  and
      Kozareva, Zornitsa  and
      Zhang, Yue",
    booktitle = "Proceedings of the 2022 Conference on Empirical Methods in Natural Language Processing",
    month = dec,
    year = "2022",
    address = "Abu Dhabi, United Arab Emirates",
    publisher = "Association for Computational Linguistics",
    url = "https://aclanthology.org/2022.emnlp-main.14/",
    doi = "10.18653/v1/2022.emnlp-main.14",
    pages = "184--198"
}

@inproceedings{jumelet-etal-2021-language,
    title = "Language Models Use Monotonicity to Assess {NPI} Licensing",
    author = "Jumelet, Jaap  and
      Denic, Milica  and
      Szymanik, Jakub  and
      Hupkes, Dieuwke  and
      Steinert-Threlkeld, Shane",
    editor = "Zong, Chengqing  and
      Xia, Fei  and
      Li, Wenjie  and
      Navigli, Roberto",
    booktitle = "Findings of the Association for Computational Linguistics: ACL-IJCNLP 2021",
    month = aug,
    year = "2021",
    address = "Online",
    publisher = "Association for Computational Linguistics",
    url = "https://aclanthology.org/2021.findings-acl.439/",
    doi = "10.18653/v1/2021.findings-acl.439",
    pages = "4958--4969"
}

@inproceedings{DBLP:conf/iclr/WalLMSSZB25,
  author       = {Oskar van der Wal and
                  Pietro Lesci and
                  Max M{\"{u}}ller{-}Eberstein and
                  Naomi Saphra and
                  Hailey Schoelkopf and
                  Willem H. Zuidema and
                  Stella Biderman},
  title        = {PolyPythias: Stability and Outliers across Fifty Language Model Pre-Training
                  Runs},
  booktitle    = {The Thirteenth International Conference on Learning Representations,
                  {ICLR} 2025, Singapore, April 24-28, 2025},
  publisher    = {OpenReview.net},
  year         = {2025},
  url          = {https://openreview.net/forum?id=bmrYu2Ekdz},
  bibsource    = {dblp computer science bibliography, https://dblp.org}
}

@article{DBLP:journals/corr/abs-2510-24963,
  author       = {James A. Michaelov and
                  Roger P. Levy and
                  Benjamin K. Bergen},
  title        = {Language Model Behavioral Phases are Consistent Across Architecture,
                  Training Data, and Scale},
  journal      = {CoRR},
  volume       = {abs/2510.24963},
  year         = {2025},
  url          = {https://doi.org/10.48550/arXiv.2510.24963},
  doi          = {10.48550/ARXIV.2510.24963},
  eprinttype    = {arXiv},
  eprint       = {2510.24963},
  bibsource    = {dblp computer science bibliography, https://dblp.org}
}

@article{honnibal2020spacy,
  author = {Honnibal, Matthew and Montani, Ines and Van Landeghem, Sofie and Boyd, Adriane},
  doi = {10.5281/zenodo.1212303},
  title = {{spaCy: Industrial-strength Natural Language Processing in Python}},
  year = 2020
}

@inproceedings{
liu2024infinigram,
title={Infini-gram: Scaling Unbounded n-gram Language Models to a Trillion Tokens},
author={Jiacheng Liu and Sewon Min and Luke Zettlemoyer and Yejin Choi and Hannaneh Hajishirzi},
booktitle={First Conference on Language Modeling},
year={2024},
url={https://openreview.net/forum?id=u2vAyMeLMm}
}

@article{jumelet2025misra,
  author       = {Jaap Jumelet and
                  Kanishka Misra and
                  Lisa Bylinina and
                  Jakub Szymanik and
                  Jennifer Culbertson and
                  Kyle Mahowald},
  title        = {Uncovering the Pathways to Word Order Acquisition in Language Models},
  journal      = {Under Review},
  year         = {2025},
}

@book{diessel2019grammar,
  title={The grammar network: How linguistic structure is shaped by language use.},
  author={Diessel, Holger},
  year={2019},
  publisher={Cambridge University Press}
}

@article{macdonald2013language,
  title={How language production shapes language form and comprehension},
  author={MacDonald, Maryellen C},
  journal={Frontiers in psychology},
  volume={4},
  pages={226},
  year={2013},
  publisher={Frontiers Media SA}
}

@inproceedings{teodorescu2006adjective,
  title={Adjective Ordering Restrictions Revisited},
  author={Teodorescu, Alexandra},
  booktitle={25th West Coast Conference on Formal Linguistics},
  pages={399--407},
  year={2006},
  organization={Cascadilla Proceedings Project}
}

@article{dyer-etal-2023-evaluating,
    title = "Evaluating a Century of Progress on the Cognitive Science of Adjective Ordering",
    author = "Dyer, William  and
      Torres, Charles  and
      Scontras, Gregory  and
      Futrell, Richard",
    journal = "Transactions of the Association for Computational Linguistics",
    volume = "11",
    year = "2023",
    address = "Cambridge, MA",
    publisher = "MIT Press",
    url = "https://aclanthology.org/2023.tacl-1.67",
    doi = "10.1162/tacl_a_00596",
    pages = "1185--1200",
}

@inproceedings{dyer-etal-2021-predicting,
    title = "Predicting cross-linguistic adjective order with information gain",
    author = "Dyer, William  and
      Futrell, Richard  and
      Liu, Zoey  and
      Scontras, Greg",
    editor = "Zong, Chengqing  and
      Xia, Fei  and
      Li, Wenjie  and
      Navigli, Roberto",
    booktitle = "Findings of the Association for Computational Linguistics: ACL-IJCNLP 2021",
    month = aug,
    year = "2021",
    address = "Online",
    publisher = "Association for Computational Linguistics",
    url = "https://aclanthology.org/2021.findings-acl.83",
    doi = "10.18653/v1/2021.findings-acl.83",
    pages = "957--967",
}

@inproceedings{futrell-etal-2020-determines,
    title = "What determines the order of adjectives in {E}nglish? Comparing efficiency-based theories using dependency treebanks",
    author = "Futrell, Richard  and
      Dyer, William  and
      Scontras, Greg",
    editor = "Jurafsky, Dan  and
      Chai, Joyce  and
      Schluter, Natalie  and
      Tetreault, Joel",
    booktitle = "Proceedings of the 58th Annual Meeting of the Association for Computational Linguistics",
    month = jul,
    year = "2020",
    address = "Online",
    publisher = "Association for Computational Linguistics",
    url = "https://aclanthology.org/2020.acl-main.181",
    doi = "10.18653/v1/2020.acl-main.181",
    pages = "2003--2012",
}

@article{scontras2017,
    author = {Scontras, Gregory and Degen, Judith and Goodman, Noah D.},
    title = "{Subjectivity Predicts Adjective Ordering Preferences}",
    journal = {Open Mind},
    volume = {1},
    number = {1},
    pages = {53-66},
    year = {2017},
    month = {02},
    issn = {2470-2986},
    doi = {10.1162/OPMI_a_00005},
    url = {https://doi.org/10.1162/OPMI\_a\_00005},
    eprint = {https://direct.mit.edu/opmi/article-pdf/1/1/53/1868262/opmi\_a\_00005.pdf},
}

@article{ney1983,
    author = {James W. Ney},
    title = {Optionality and choice in the selection of verb complements in English},
    journal = {WORD},
    volume = {32},
    number = {2},
    pages = {133--152},
    year = {1981},
    publisher = {Routledge},
    doi = {10.1080/00437956.1981.11435707},
    URL = {https://doi.org/10.1080/00437956.1981.11435707},
    eprint = {https://doi.org/10.1080/00437956.1981.11435707}
}

@article{behaghel_von_1930,
	title = {Von deutscher {Wortstellung}},
	volume = {44},
	journal = {Zeitschrift für Deutschkunde},
	author = {Behaghel, Otto},
	year = {1930},
	pages = {81--89}
}

@article{GIBSON19981,
title = {Linguistic complexity: locality of syntactic dependencies},
journal = {Cognition},
volume = {68},
number = {1},
pages = {1-76},
year = {1998},
issn = {0010-0277},
doi = {https://doi.org/10.1016/S0010-0277(98)00034-1},
url = {https://www.sciencedirect.com/science/article/pii/S0010027798000341},
author = {Edward Gibson},
}

@article{Hahn2018AnIE,
  title={An Information-Theoretic Explanation of Adjective Ordering Preferences},
  author={Michael Hahn and Judith Degen and Noah D. Goodman and Dan Jurafsky and Richard Futrell},
  journal={Cognitive Science},
  year={2018},
  url={https://api.semanticscholar.org/CorpusID:21719450}
}

@article{wulff2015,
author = {Wulff, Stefanie and Gries, Stefan},
year = {2015},
month = {05},
pages = {},
title = {Prenominal adjective order preferences in Chinese and German L2 English: A multifactorial corpus study},
volume = {5},
journal = {Linguistic Approaches to Bilingualism},
doi = {10.1075/lab.5.1.05wul}
}

@article{warstadt-etal-2020-blimp-benchmark,
    title = "{BL}i{MP}: The Benchmark of Linguistic Minimal Pairs for {E}nglish",
    author = "Warstadt, Alex  and
      Parrish, Alicia  and
      Liu, Haokun  and
      Mohananey, Anhad  and
      Peng, Wei  and
      Wang, Sheng-Fu  and
      Bowman, Samuel R.",
    editor = "Johnson, Mark  and
      Roark, Brian  and
      Nenkova, Ani",
    journal = "Transactions of the Association for Computational Linguistics",
    volume = "8",
    year = "2020",
    address = "Cambridge, MA",
    publisher = "MIT Press",
    url = "https://aclanthology.org/2020.tacl-1.25",
    doi = "10.1162/tacl_a_00321",
    pages = "377--392",
}

@article{Futrell2020DependencyLA,
  title={Dependency locality as an explanatory principle for word order},
  author={Richard Futrell and Roger Philip Levy and Edward Gibson},
  journal={Language},
  year={2020},
  volume={96},
  pages={371 - 412},
  url={https://api.semanticscholar.org/CorpusID:220495365}
}

@inproceedings{wilcox-etal-2018-rnn,
    title = "What do {RNN} Language Models Learn about Filler{--}Gap Dependencies?",
    author = "Wilcox, Ethan  and
      Levy, Roger  and
      Morita, Takashi  and
      Futrell, Richard",
    editor = "Linzen, Tal  and
      Chrupa{\l}a, Grzegorz  and
      Alishahi, Afra",
    booktitle = "Proceedings of the 2018 {EMNLP} Workshop {B}lackbox{NLP}: Analyzing and Interpreting Neural Networks for {NLP}",
    month = nov,
    year = "2018",
    address = "Brussels, Belgium",
    publisher = "Association for Computational Linguistics",
    url = "https://aclanthology.org/W18-5423",
    doi = "10.18653/v1/W18-5423",
    pages = "211--221",
}

@inproceedings{suijkerbuijk-etal-2023-learnability,
    title = "The Learnability of the Wh-Island Constraint in {D}utch by a Long Short-Term Memory Network",
    author = "Suijkerbuijk, Michelle  and
      de Swart, Peter  and
      Frank, Stefan L.",
    editor = "Hunter, Tim  and
      Prickett, Brandon",
    booktitle = "Proceedings of the Society for Computation in Linguistics 2023",
    month = jun,
    year = "2023",
    address = "Amherst, MA",
    publisher = "Association for Computational Linguistics",
    url = "https://aclanthology.org/2023.scil-1.28",
    pages = "321--331",
}

@inproceedings{gulordava-etal-2018-colorless,
    title = "Colorless Green Recurrent Networks Dream Hierarchically",
    author = "Gulordava, Kristina  and
      Bojanowski, Piotr  and
      Grave, Edouard  and
      Linzen, Tal  and
      Baroni, Marco",
    editor = "Walker, Marilyn  and
      Ji, Heng  and
      Stent, Amanda",
    booktitle = "Proceedings of the 2018 Conference of the North {A}merican Chapter of the Association for Computational Linguistics: Human Language Technologies, Volume 1 (Long Papers)",
    month = jun,
    year = "2018",
    address = "New Orleans, Louisiana",
    publisher = "Association for Computational Linguistics",
    url = "https://aclanthology.org/N18-1108",
    doi = "10.18653/v1/N18-1108",
    pages = "1195--1205",
}

@inproceedings{bylinina-tikhonov-2022-transformers,
    title = "Transformers in the loop: Polarity in neural models of language",
    author = "Bylinina, Lisa  and
      Tikhonov, Alexey",
    editor = "Muresan, Smaranda  and
      Nakov, Preslav  and
      Villavicencio, Aline",
    booktitle = "Proceedings of the 60th Annual Meeting of the Association for Computational Linguistics (Volume 1: Long Papers)",
    month = may,
    year = "2022",
    address = "Dublin, Ireland",
    publisher = "Association for Computational Linguistics",
    url = "https://aclanthology.org/2022.acl-long.455",
    doi = "10.18653/v1/2022.acl-long.455",
    pages = "6601--6610",
}

@article{linzen-etal-2016-assessing,
    title = "Assessing the Ability of {LSTM}s to Learn Syntax-Sensitive Dependencies",
    author = "Linzen, Tal  and
      Dupoux, Emmanuel  and
      Goldberg, Yoav",
    editor = "Lee, Lillian  and
      Johnson, Mark  and
      Toutanova, Kristina",
    journal = "Transactions of the Association for Computational Linguistics",
    volume = "4",
    year = "2016",
    address = "Cambridge, MA",
    publisher = "MIT Press",
    url = "https://aclanthology.org/Q16-1037",
    doi = "10.1162/tacl_a_00115",
    pages = "521--535",
}

@inproceedings{jumelet-hupkes-2018-language,
    title = "Do Language Models Understand Anything? On the Ability of {LSTM}s to Understand Negative Polarity Items",
    author = "Jumelet, Jaap  and
      Hupkes, Dieuwke",
    editor = "Linzen, Tal  and
      Chrupa{\l}a, Grzegorz  and
      Alishahi, Afra",
    booktitle = "Proceedings of the 2018 {EMNLP} Workshop {B}lackbox{NLP}: Analyzing and Interpreting Neural Networks for {NLP}",
    month = nov,
    year = "2018",
    address = "Brussels, Belgium",
    publisher = "Association for Computational Linguistics",
    url = "https://aclanthology.org/W18-5424",
    doi = "10.18653/v1/W18-5424",
    pages = "222--231",
}

@inproceedings{gauthier-etal-2020-syntaxgym,
    title = "{S}yntax{G}ym: An Online Platform for Targeted Evaluation of Language Models",
    author = "Gauthier, Jon  and
      Hu, Jennifer  and
      Wilcox, Ethan  and
      Qian, Peng  and
      Levy, Roger",
    editor = "Celikyilmaz, Asli  and
      Wen, Tsung-Hsien",
    booktitle = "Proceedings of the 58th Annual Meeting of the Association for Computational Linguistics: System Demonstrations",
    month = jul,
    year = "2020",
    address = "Online",
    publisher = "Association for Computational Linguistics",
    url = "https://aclanthology.org/2020.acl-demos.10",
    doi = "10.18653/v1/2020.acl-demos.10",
    pages = "70--76",
}

@inproceedings{marvin-linzen-2018-targeted,
    title = "Targeted Syntactic Evaluation of Language Models",
    author = "Marvin, Rebecca  and
      Linzen, Tal",
    editor = "Riloff, Ellen  and
      Chiang, David  and
      Hockenmaier, Julia  and
      Tsujii, Jun{'}ichi",
    booktitle = "Proceedings of the 2018 Conference on Empirical Methods in Natural Language Processing",
    month = oct # "-" # nov,
    year = "2018",
    address = "Brussels, Belgium",
    publisher = "Association for Computational Linguistics",
    url = "https://aclanthology.org/D18-1151",
    doi = "10.18653/v1/D18-1151",
    pages = "1192--1202",
}

@article{linzen2021baroni,
   author = "Linzen, Tal and Baroni, Marco",
   title = "Syntactic Structure from Deep Learning", 
   journal= "Annual Review of Linguistics",
   year = "2021",
   volume = "7",
   number = "Volume 7, 2021",
   pages = "195-212",
   doi = "https://doi.org/10.1146/annurev-linguistics-032020-051035",
   url = "https://www.annualreviews.org/content/journals/10.1146/annurev-linguistics-032020-051035",
   publisher = "Annual Reviews",
   issn = "2333-9691",
   type = "Journal Article",
  }

@article{scontras2023adjective,
  title={Adjective ordering across languages},
  author={Scontras, Gregory},
  journal={Annual Review of Linguistics},
  volume={9},
  pages={357--376},
  year={2023},
  publisher={Annual Reviews}
}

@inproceedings{hill-2012-beauty,
    title = "Beauty Before Age? {A}pplying Subjectivity to Automatic {E}nglish Adjective Ordering",
    author = "Hill, Felix",
    editor = "Levitan, Rivka  and
      Ott, Myle  and
      Levy, Roger  and
      Nenkova, Ani",
    booktitle = "Proceedings of the {NAACL} {HLT} 2012 Student Research Workshop",
    month = jun,
    year = "2012",
    address = "Montr{\'e}al, Canada",
    publisher = "Association for Computational Linguistics",
    url = "https://aclanthology.org/N12-2003",
    pages = "11--16",
}

@inproceedings{leung-etal-2020-investigating,
    title = "Investigating {C}ross-{L}inguistic {A}djective {O}rdering {T}endencies with a {L}atent-{V}ariable {M}odel",
    author = "Leung, Jun Yen  and
      Emerson, Guy  and
      Cotterell, Ryan",
    editor = "Webber, Bonnie  and
      Cohn, Trevor  and
      He, Yulan  and
      Liu, Yang",
    booktitle = "Proceedings of the 2020 Conference on Empirical Methods in Natural Language Processing (EMNLP)",
    month = nov,
    year = "2020",
    address = "Online",
    publisher = "Association for Computational Linguistics",
    url = "https://aclanthology.org/2020.emnlp-main.329",
    doi = "10.18653/v1/2020.emnlp-main.329",
    pages = "4016--4028",
}

@article{BYRNE197973,
title = {Rules of prenominal adjective order and the interpretation of “incompatible” adjective pairs},
journal = {Journal of Verbal Learning and Verbal Behavior},
volume = {18},
number = {1},
pages = {73-78},
year = {1979},
issn = {0022-5371},
doi = {https://doi.org/10.1016/S0022-5371(79)90574-7},
url = {https://www.sciencedirect.com/science/article/pii/S0022537179905747},
author = {Brian Byrne}
}

@article{FERREIRA2000296,
title = {Effect of Ambiguity and Lexical Availability on Syntactic and Lexical Production},
journal = {Cognitive Psychology},
volume = {40},
number = {4},
pages = {296-340},
year = {2000},
issn = {0010-0285},
doi = {https://doi.org/10.1006/cogp.1999.0730},
url = {https://www.sciencedirect.com/science/article/pii/S0010028599907302},
author = {Victor S. Ferreira and Gary S. Dell}
}

@misc{yang2025qwen3technicalreport,
      title={Qwen3 Technical Report}, 
      author={An Yang and Anfeng Li and Baosong Yang and Beichen Zhang and Binyuan Hui and Bo Zheng and Bowen Yu and Chang Gao and Chengen Huang and Chenxu Lv and Chujie Zheng and Dayiheng Liu and Fan Zhou and Fei Huang and Feng Hu and Hao Ge and Haoran Wei and Huan Lin and Jialong Tang and Jian Yang and Jianhong Tu and Jianwei Zhang and Jianxin Yang and Jiaxi Yang and Jing Zhou and Jingren Zhou and Junyang Lin and Kai Dang and Keqin Bao and Kexin Yang and Le Yu and Lianghao Deng and Mei Li and Mingfeng Xue and Mingze Li and Pei Zhang and Peng Wang and Qin Zhu and Rui Men and Ruize Gao and Shixuan Liu and Shuang Luo and Tianhao Li and Tianyi Tang and Wenbiao Yin and Xingzhang Ren and Xinyu Wang and Xinyu Zhang and Xuancheng Ren and Yang Fan and Yang Su and Yichang Zhang and Yinger Zhang and Yu Wan and Yuqiong Liu and Zekun Wang and Zeyu Cui and Zhenru Zhang and Zhipeng Zhou and Zihan Qiu},
      year={2025},
      eprint={2505.09388},
      archivePrefix={arXiv},
      primaryClass={cs.CL},
      url={https://arxiv.org/abs/2505.09388}, 
}

@misc{grattafiori2024llama3herdmodels,
      title={The Llama 3 Herd of Models}, 
      author={Aaron Grattafiori and Abhimanyu Dubey and Abhinav Jauhri and Abhinav Pandey and Abhishek Kadian and Ahmad Al-Dahle and Aiesha Letman and Akhil Mathur and Alan Schelten and Alex Vaughan and Amy Yang and Angela Fan and Anirudh Goyal and Anthony Hartshorn and Aobo Yang and Archi Mitra and Archie Sravankumar and Artem Korenev and Arthur Hinsvark and Arun Rao and Aston Zhang and Aurelien Rodriguez and Austen Gregerson and Ava Spataru and Baptiste Roziere and Bethany Biron and Binh Tang and Bobbie Chern and Charlotte Caucheteux and Chaya Nayak and Chloe Bi and Chris Marra and Chris McConnell and Christian Keller and Christophe Touret and Chunyang Wu and Corinne Wong and Cristian Canton Ferrer and Cyrus Nikolaidis and Damien Allonsius and Daniel Song and Danielle Pintz and Danny Livshits and Danny Wyatt and David Esiobu and Dhruv Choudhary and Dhruv Mahajan and Diego Garcia-Olano and Diego Perino and Dieuwke Hupkes and Egor Lakomkin and Ehab AlBadawy and Elina Lobanova and Emily Dinan and Eric Michael Smith and Filip Radenovic and Francisco Guzmán and Frank Zhang and Gabriel Synnaeve and Gabrielle Lee and Georgia Lewis Anderson and Govind Thattai and Graeme Nail and Gregoire Mialon and Guan Pang and Guillem Cucurell and Hailey Nguyen and Hannah Korevaar and Hu Xu and Hugo Touvron and Iliyan Zarov and Imanol Arrieta Ibarra and Isabel Kloumann and Ishan Misra and Ivan Evtimov and Jack Zhang and Jade Copet and Jaewon Lee and Jan Geffert and Jana Vranes and Jason Park and Jay Mahadeokar and Jeet Shah and Jelmer van der Linde and Jennifer Billock and Jenny Hong and Jenya Lee and Jeremy Fu and Jianfeng Chi and Jianyu Huang and Jiawen Liu and Jie Wang and Jiecao Yu and Joanna Bitton and Joe Spisak and Jongsoo Park and Joseph Rocca and Joshua Johnstun and Joshua Saxe and Junteng Jia and Kalyan Vasuden Alwala and Karthik Prasad and Kartikeya Upasani and Kate Plawiak and Ke Li and Kenneth Heafield and Kevin Stone and Khalid El-Arini and Krithika Iyer and Kshitiz Malik and Kuenley Chiu and Kunal Bhalla and Kushal Lakhotia and Lauren Rantala-Yeary and Laurens van der Maaten and Lawrence Chen and Liang Tan and Liz Jenkins and Louis Martin and Lovish Madaan and Lubo Malo and Lukas Blecher and Lukas Landzaat and Luke de Oliveira and Madeline Muzzi and Mahesh Pasupuleti and Mannat Singh and Manohar Paluri and Marcin Kardas and Maria Tsimpoukelli and Mathew Oldham and Mathieu Rita and Maya Pavlova and Melanie Kambadur and Mike Lewis and Min Si and Mitesh Kumar Singh and Mona Hassan and Naman Goyal and Narjes Torabi and Nikolay Bashlykov and Nikolay Bogoychev and Niladri Chatterji and Ning Zhang and Olivier Duchenne and Onur Çelebi and Patrick Alrassy and Pengchuan Zhang and Pengwei Li and Petar Vasic and Peter Weng and Prajjwal Bhargava and Pratik Dubal and Praveen Krishnan and Punit Singh Koura and Puxin Xu and Qing He and Qingxiao Dong and Ragavan Srinivasan and Raj Ganapathy and Ramon Calderer and Ricardo Silveira Cabral and Robert Stojnic and Roberta Raileanu and Rohan Maheswari and Rohit Girdhar and Rohit Patel and Romain Sauvestre and Ronnie Polidoro and Roshan Sumbaly and Ross Taylor and Ruan Silva and Rui Hou and Rui Wang and Saghar Hosseini and Sahana Chennabasappa and Sanjay Singh and Sean Bell and Seohyun Sonia Kim and Sergey Edunov and Shaoliang Nie and Sharan Narang and Sharath Raparthy and Sheng Shen and Shengye Wan and Shruti Bhosale and Shun Zhang and Simon Vandenhende and Soumya Batra and Spencer Whitman and Sten Sootla and Stephane Collot and Suchin Gururangan and Sydney Borodinsky and Tamar Herman and Tara Fowler and Tarek Sheasha and Thomas Georgiou and Thomas Scialom and Tobias Speckbacher and Todor Mihaylov and Tong Xiao and Ujjwal Karn and Vedanuj Goswami and Vibhor Gupta and Vignesh Ramanathan and Viktor Kerkez and Vincent Gonguet and Virginie Do and Vish Vogeti and Vítor Albiero and Vladan Petrovic and Weiwei Chu and Wenhan Xiong and Wenyin Fu and Whitney Meers and Xavier Martinet and Xiaodong Wang and Xiaofang Wang and Xiaoqing Ellen Tan and Xide Xia and Xinfeng Xie and Xuchao Jia and Xuewei Wang and Yaelle Goldschlag and Yashesh Gaur and Yasmine Babaei and Yi Wen and Yiwen Song and Yuchen Zhang and Yue Li and Yuning Mao and Zacharie Delpierre Coudert and Zheng Yan and Zhengxing Chen and Zoe Papakipos and Aaditya Singh and Aayushi Srivastava and Abha Jain and Adam Kelsey and Adam Shajnfeld and Adithya Gangidi and Adolfo Victoria and Ahuva Goldstand and Ajay Menon and Ajay Sharma and Alex Boesenberg and Alexei Baevski and Allie Feinstein and Amanda Kallet and Amit Sangani and Amos Teo and Anam Yunus and Andrei Lupu and Andres Alvarado and Andrew Caples and Andrew Gu and Andrew Ho and Andrew Poulton and Andrew Ryan and Ankit Ramchandani and Annie Dong and Annie Franco and Anuj Goyal and Aparajita Saraf and Arkabandhu Chowdhury and Ashley Gabriel and Ashwin Bharambe and Assaf Eisenman and Azadeh Yazdan and Beau James and Ben Maurer and Benjamin Leonhardi and Bernie Huang and Beth Loyd and Beto De Paola and Bhargavi Paranjape and Bing Liu and Bo Wu and Boyu Ni and Braden Hancock and Bram Wasti and Brandon Spence and Brani Stojkovic and Brian Gamido and Britt Montalvo and Carl Parker and Carly Burton and Catalina Mejia and Ce Liu and Changhan Wang and Changkyu Kim and Chao Zhou and Chester Hu and Ching-Hsiang Chu and Chris Cai and Chris Tindal and Christoph Feichtenhofer and Cynthia Gao and Damon Civin and Dana Beaty and Daniel Kreymer and Daniel Li and David Adkins and David Xu and Davide Testuggine and Delia David and Devi Parikh and Diana Liskovich and Didem Foss and Dingkang Wang and Duc Le and Dustin Holland and Edward Dowling and Eissa Jamil and Elaine Montgomery and Eleonora Presani and Emily Hahn and Emily Wood and Eric-Tuan Le and Erik Brinkman and Esteban Arcaute and Evan Dunbar and Evan Smothers and Fei Sun and Felix Kreuk and Feng Tian and Filippos Kokkinos and Firat Ozgenel and Francesco Caggioni and Frank Kanayet and Frank Seide and Gabriela Medina Florez and Gabriella Schwarz and Gada Badeer and Georgia Swee and Gil Halpern and Grant Herman and Grigory Sizov and Guangyi and Zhang and Guna Lakshminarayanan and Hakan Inan and Hamid Shojanazeri and Han Zou and Hannah Wang and Hanwen Zha and Haroun Habeeb and Harrison Rudolph and Helen Suk and Henry Aspegren and Hunter Goldman and Hongyuan Zhan and Ibrahim Damlaj and Igor Molybog and Igor Tufanov and Ilias Leontiadis and Irina-Elena Veliche and Itai Gat and Jake Weissman and James Geboski and James Kohli and Janice Lam and Japhet Asher and Jean-Baptiste Gaya and Jeff Marcus and Jeff Tang and Jennifer Chan and Jenny Zhen and Jeremy Reizenstein and Jeremy Teboul and Jessica Zhong and Jian Jin and Jingyi Yang and Joe Cummings and Jon Carvill and Jon Shepard and Jonathan McPhie and Jonathan Torres and Josh Ginsburg and Junjie Wang and Kai Wu and Kam Hou U and Karan Saxena and Kartikay Khandelwal and Katayoun Zand and Kathy Matosich and Kaushik Veeraraghavan and Kelly Michelena and Keqian Li and Kiran Jagadeesh and Kun Huang and Kunal Chawla and Kyle Huang and Lailin Chen and Lakshya Garg and Lavender A and Leandro Silva and Lee Bell and Lei Zhang and Liangpeng Guo and Licheng Yu and Liron Moshkovich and Luca Wehrstedt and Madian Khabsa and Manav Avalani and Manish Bhatt and Martynas Mankus and Matan Hasson and Matthew Lennie and Matthias Reso and Maxim Groshev and Maxim Naumov and Maya Lathi and Meghan Keneally and Miao Liu and Michael L. Seltzer and Michal Valko and Michelle Restrepo and Mihir Patel and Mik Vyatskov and Mikayel Samvelyan and Mike Clark and Mike Macey and Mike Wang and Miquel Jubert Hermoso and Mo Metanat and Mohammad Rastegari and Munish Bansal and Nandhini Santhanam and Natascha Parks and Natasha White and Navyata Bawa and Nayan Singhal and Nick Egebo and Nicolas Usunier and Nikhil Mehta and Nikolay Pavlovich Laptev and Ning Dong and Norman Cheng and Oleg Chernoguz and Olivia Hart and Omkar Salpekar and Ozlem Kalinli and Parkin Kent and Parth Parekh and Paul Saab and Pavan Balaji and Pedro Rittner and Philip Bontrager and Pierre Roux and Piotr Dollar and Polina Zvyagina and Prashant Ratanchandani and Pritish Yuvraj and Qian Liang and Rachad Alao and Rachel Rodriguez and Rafi Ayub and Raghotham Murthy and Raghu Nayani and Rahul Mitra and Rangaprabhu Parthasarathy and Raymond Li and Rebekkah Hogan and Robin Battey and Rocky Wang and Russ Howes and Ruty Rinott and Sachin Mehta and Sachin Siby and Sai Jayesh Bondu and Samyak Datta and Sara Chugh and Sara Hunt and Sargun Dhillon and Sasha Sidorov and Satadru Pan and Saurabh Mahajan and Saurabh Verma and Seiji Yamamoto and Sharadh Ramaswamy and Shaun Lindsay and Shaun Lindsay and Sheng Feng and Shenghao Lin and Shengxin Cindy Zha and Shishir Patil and Shiva Shankar and Shuqiang Zhang and Shuqiang Zhang and Sinong Wang and Sneha Agarwal and Soji Sajuyigbe and Soumith Chintala and Stephanie Max and Stephen Chen and Steve Kehoe and Steve Satterfield and Sudarshan Govindaprasad and Sumit Gupta and Summer Deng and Sungmin Cho and Sunny Virk and Suraj Subramanian and Sy Choudhury and Sydney Goldman and Tal Remez and Tamar Glaser and Tamara Best and Thilo Koehler and Thomas Robinson and Tianhe Li and Tianjun Zhang and Tim Matthews and Timothy Chou and Tzook Shaked and Varun Vontimitta and Victoria Ajayi and Victoria Montanez and Vijai Mohan and Vinay Satish Kumar and Vishal Mangla and Vlad Ionescu and Vlad Poenaru and Vlad Tiberiu Mihailescu and Vladimir Ivanov and Wei Li and Wenchen Wang and Wenwen Jiang and Wes Bouaziz and Will Constable and Xiaocheng Tang and Xiaojian Wu and Xiaolan Wang and Xilun Wu and Xinbo Gao and Yaniv Kleinman and Yanjun Chen and Ye Hu and Ye Jia and Ye Qi and Yenda Li and Yilin Zhang and Ying Zhang and Yossi Adi and Youngjin Nam and Yu and Wang and Yu Zhao and Yuchen Hao and Yundi Qian and Yunlu Li and Yuzi He and Zach Rait and Zachary DeVito and Zef Rosnbrick and Zhaoduo Wen and Zhenyu Yang and Zhiwei Zhao and Zhiyu Ma},
      year={2024},
      eprint={2407.21783},
      archivePrefix={arXiv},
      primaryClass={cs.AI},
      url={https://arxiv.org/abs/2407.21783}, 
}

@misc{olmo2025olmo3,
      title={Olmo 3}, 
      author={Team OLMo and : and Allyson Ettinger and Amanda Bertsch and Bailey Kuehl and David Graham and David Heineman and Dirk Groeneveld and Faeze Brahman and Finbarr Timbers and Hamish Ivison and Jacob Morrison and Jake Poznanski and Kyle Lo and Luca Soldaini and Matt Jordan and Mayee Chen and Michael Noukhovitch and Nathan Lambert and Pete Walsh and Pradeep Dasigi and Robert Berry and Saumya Malik and Saurabh Shah and Scott Geng and Shane Arora and Shashank Gupta and Taira Anderson and Teng Xiao and Tyler Murray and Tyler Romero and Victoria Graf and Akari Asai and Akshita Bhagia and Alexander Wettig and Alisa Liu and Aman Rangapur and Chloe Anastasiades and Costa Huang and Dustin Schwenk and Harsh Trivedi and Ian Magnusson and Jaron Lochner and Jiacheng Liu and Lester James V. Miranda and Maarten Sap and Malia Morgan and Michael Schmitz and Michal Guerquin and Michael Wilson and Regan Huff and Ronan Le Bras and Rui Xin and Rulin Shao and Sam Skjonsberg and Shannon Zejiang Shen and Shuyue Stella Li and Tucker Wilde and Valentina Pyatkin and Will Merrill and Yapei Chang and Yuling Gu and Zhiyuan Zeng and Ashish Sabharwal and Luke Zettlemoyer and Pang Wei Koh and Ali Farhadi and Noah A. Smith and Hannaneh Hajishirzi},
      year={2025},
      eprint={2512.13961},
      archivePrefix={arXiv},
      primaryClass={cs.CL},
      url={https://arxiv.org/abs/2512.13961}, 
}

@misc{olmo20252olmo2furious,
      title={2 OLMo 2 Furious}, 
      author={Team OLMo and Pete Walsh and Luca Soldaini and Dirk Groeneveld and Kyle Lo and Shane Arora and Akshita Bhagia and Yuling Gu and Shengyi Huang and Matt Jordan and Nathan Lambert and Dustin Schwenk and Oyvind Tafjord and Taira Anderson and David Atkinson and Faeze Brahman and Christopher Clark and Pradeep Dasigi and Nouha Dziri and Allyson Ettinger and Michal Guerquin and David Heineman and Hamish Ivison and Pang Wei Koh and Jiacheng Liu and Saumya Malik and William Merrill and Lester James V. Miranda and Jacob Morrison and Tyler Murray and Crystal Nam and Jake Poznanski and Valentina Pyatkin and Aman Rangapur and Michael Schmitz and Sam Skjonsberg and David Wadden and Christopher Wilhelm and Michael Wilson and Luke Zettlemoyer and Ali Farhadi and Noah A. Smith and Hannaneh Hajishirzi},
      year={2025},
      eprint={2501.00656},
      archivePrefix={arXiv},
      primaryClass={cs.CL},
      url={https://arxiv.org/abs/2501.00656}, 
}

@inproceedings{DBLP:conf/hhai/NeelySBL22,
  author       = {Michael Neely and
                  Stefan F. Schouten and
                  Maurits J. R. Bleeker and
                  Ana Lucic},
  editor       = {Stefan Schlobach and
                  Mar{\'{\i}}a P{\'{e}}rez{-}Ortiz and
                  Myrthe Tielman},
  title        = {A Song of (Dis)agreement: Evaluating the Evaluation of Explainable
                  Artificial Intelligence in Natural Language Processing},
  booktitle    = {{HHAI} 2022: Augmenting Human Intellect - Proceedings of the First
                  International Conference on Hybrid Human-Artificial Intelligence,
                  Amsterdam, The Netherlands, 13-17 June 2022},
  series       = {Frontiers in Artificial Intelligence and Applications},
  volume       = {354},
  pages        = {60--78},
  publisher    = {{IOS} Press},
  year         = {2022},
  url          = {https://doi.org/10.3233/FAIA220190},
  doi          = {10.3233/FAIA220190},
  bibsource    = {dblp computer science bibliography, https://dblp.org}
}

@article{10.1162/tacl_a_00444,
    author = {Chang, Tyler A. and Bergen, Benjamin K.},
    title = {Word Acquisition in Neural Language Models},
    journal = {Transactions of the Association for Computational Linguistics},
    volume = {10},
    pages = {1-16},
    year = {2022},
    month = {01},
    issn = {2307-387X},
    doi = {10.1162/tacl_a_00444},
    url = {https://doi.org/10.1162/tacl_a_00444},
    eprint = {https://direct.mit.edu/tacl/article-pdf/doi/10.1162/tacl_a_00444/1986589/tacl_a_00444.pdf},
}

@inproceedings{muller-eberstein-etal-2025-decaf,
    title = "{DECAF}: A Dynamically Extensible Corpus Analysis Framework",
    author = {M{\"u}ller-Eberstein, Max  and
      Goot, Rob Van Der  and
      Rogers, Anna},
    editor = "Mishra, Pushkar  and
      Muresan, Smaranda  and
      Yu, Tao",
    booktitle = "Proceedings of the 63rd Annual Meeting of the Association for Computational Linguistics (Volume 3: System Demonstrations)",
    month = jul,
    year = "2025",
    address = "Vienna, Austria",
    publisher = "Association for Computational Linguistics",
    url = "https://aclanthology.org/2025.acl-demo.34/",
    doi = "10.18653/v1/2025.acl-demo.34",
    pages = "351--362",
    ISBN = "979-8-89176-253-4"
}

@article{culbertson2020,
author = {Culbertson, Jennifer and Schouwstra, Marieke and Kirby, Simon},
year = {2020},
month = {01},
pages = {696-717},
title = {From the world to word order: Deriving biases in noun phrase order from statistical properties of the world},
volume = {96},
journal = {Language},
doi = {10.1353/lan.2020.0045}
}

@article{Futrell_2025,
   title={How Linguistics Learned to Stop Worrying and Love the Language Models},
   ISSN={1469-1825},
   url={http://dx.doi.org/10.1017/S0140525X2510112X},
   DOI={10.1017/s0140525x2510112x},
   journal={Behavioral and Brain Sciences},
   publisher={Cambridge University Press (CUP)},
   author={Futrell, Richard and Mahowald, Kyle},
   year={2025},
   month=jul, pages={1–98} }

@inproceedings{DBLP:conf/iclr/0003AEAAZW25,
  author       = {Xinyi Wang and
                  Antonis Antoniades and
                  Yanai Elazar and
                  Alfonso Amayuelas and
                  Alon Albalak and
                  Kexun Zhang and
                  William Yang Wang},
  title        = {Generalization v.s. Memorization: Tracing Language Models' Capabilities
                  Back to Pretraining Data},
  booktitle    = {The Thirteenth International Conference on Learning Representations,
                  {ICLR} 2025, Singapore, April 24-28, 2025},
  publisher    = {OpenReview.net},
  year         = {2025},
  url          = {https://openreview.net/forum?id=IQxBDLmVpT},
  bibsource    = {dblp computer science bibliography, https://dblp.org}
}

@inproceedings{lapata-etal-1999-determinants,
    title = "Determinants of Adjective-Noun Plausibility",
    author = "Lapata, Maria  and
      McDonald, Scott  and
      Keller, Frank",
    editor = "Thompson, Henry S.  and
      Lascarides, Alex",
    booktitle = "Ninth Conference of the {E}uropean Chapter of the Association for Computational Linguistics",
    month = jun,
    year = "1999",
    address = "Bergen, Norway",
    publisher = "Association for Computational Linguistics",
    url = "https://aclanthology.org/E99-1005",
    pages = "30--36",
}

@ARTICLE{10543008,
  author={Mariotti, Ettore and Arias-Duart, Anna and Cafagna, Michele and Gatt, Albert and Garcia-Gasulla, Dario and Alonso-Moral, Jose Maria},
  journal={IEEE Access}, 
  title={TextFocus: Assessing the Faithfulness of Feature Attribution Methods Explanations in Natural Language Processing}, 
  year={2024},
  volume={12},
  number={},
  pages={138870-138880},
  doi={10.1109/ACCESS.2024.3408062}}

@article{Bock1982TowardAC,
  title={Toward a Cognitive Psychology of Syntax: Information Processing Contributions to Sentence Formulation},
  author={Kathryn Bock},
  journal={Psychological Review},
  year={1982},
  volume={89},
  pages={1-47},
  url={https://api.semanticscholar.org/CorpusID:62752652}
}

@article{wulff2003,
author = {Wulff, Stefanie},
year = {2003},
month = {10},
pages = {245-282},
title = {A multifactorial corpus analysis of adjective order in English},
volume = {8},
journal = {International Journal of Corpus Linguistics},
doi = {10.1075/ijcl.8.2.04wul}
}

@misc{patil2024filtered,
      title={Filtered Corpus Training (FiCT) Shows that Language Models can Generalize from Indirect Evidence}, 
      author={Abhinav Patil and Jaap Jumelet and Yu Ying Chiu and Andy Lapastora and Peter Shen and Lexie Wang and Clevis Willrich and Shane Steinert-Threlkeld},
      year={2024},
      eprint={2405.15750},
      archivePrefix={arXiv},
      primaryClass={id='cs.CL' full_name='Computation and Language' is_active=True alt_name='cmp-lg' in_archive='cs' is_general=False description='Covers natural language processing. Roughly includes material in ACM Subject Class I.2.7. Note that work on artificial languages (programming languages, logics, formal systems) that does not explicitly address natural-language issues broadly construed (natural-language processing, computational linguistics, speech, text retrieval, etc.) is not appropriate for this area.'}
}

@book{c0f64ae7-9422-34a7-b044-fa63660e7879,
 ISBN = {9780262028844},
 URL = {http://www.jstor.org/stable/j.ctt17kk71g},
 author = {Timothy J. O’Donnell},
 publisher = {The MIT Press},
 title = {Productivity and Reuse in Language: A Theory of Linguistic Computation and Storage},
 urldate = {2024-07-02},
 year = {2015}
}

@article{DBLP:journals/corr/abs-2403-19827,
  author       = {Kanishka Misra and
                  Kyle Mahowald},
  title        = {Language Models Learn Rare Phenomena from Less Rare Phenomena: The
                  Case of the Missing AANNs},
  journal      = {CoRR},
  volume       = {abs/2403.19827},
  year         = {2024},
  url          = {https://doi.org/10.48550/arXiv.2403.19827},
  doi          = {10.48550/ARXIV.2403.19827},
  eprinttype    = {arXiv},
  eprint       = {2403.19827},
  bibsource    = {dblp computer science bibliography, https://dblp.org}
}

@misc{biderman2023pythia,
      title={Pythia: A Suite for Analyzing Large Language Models Across Training and Scaling}, 
      author={Stella Biderman and Hailey Schoelkopf and Quentin Anthony and Herbie Bradley and Kyle O'Brien and Eric Hallahan and Mohammad Aflah Khan and Shivanshu Purohit and USVSN Sai Prashanth and Edward Raff and Aviya Skowron and Lintang Sutawika and Oskar van der Wal},
      year={2023},
      eprint={2304.01373},
      archivePrefix={arXiv},
      primaryClass={cs.CL}
}

@misc{gao2020pile,
      title={The Pile: An 800GB Dataset of Diverse Text for Language Modeling}, 
      author={Leo Gao and Stella Biderman and Sid Black and Laurence Golding and Travis Hoppe and Charles Foster and Jason Phang and Horace He and Anish Thite and Noa Nabeshima and Shawn Presser and Connor Leahy},
      year={2020},
      eprint={2101.00027},
      archivePrefix={arXiv},
      primaryClass={cs.CL}
}

@inproceedings{alhama-etal-2023-linguistic,
    title = "Linguistic Productivity: the Case of Determiners in {E}nglish",
    author = "Alhama, Raquel G.  and
      Foushee, Ruthe  and
      Byrne, Daniel  and
      Ettinger, Allyson  and
      Goldin-Meadow, Susan  and
      Alishahi, Afra",
    editor = "Park, Jong C.  and
      Arase, Yuki  and
      Hu, Baotian  and
      Lu, Wei  and
      Wijaya, Derry  and
      Purwarianti, Ayu  and
      Krisnadhi, Adila Alfa",
    booktitle = "Proceedings of the 13th International Joint Conference on Natural Language Processing and the 3rd Conference of the Asia-Pacific Chapter of the Association for Computational Linguistics (Volume 1: Long Papers)",
    month = nov,
    year = "2023",
    address = "Nusa Dua, Bali",
    publisher = "Association for Computational Linguistics",
    url = "https://aclanthology.org/2023.ijcnlp-main.21",
    doi = "10.18653/v1/2023.ijcnlp-main.21",
    pages = "330--343",
}

@inproceedings{biderman2023,
 author = {Biderman, Stella and PRASHANTH, USVSN and Sutawika, Lintang and Schoelkopf, Hailey and Anthony, Quentin and Purohit, Shivanshu and Raff, Edward},
 booktitle = {Advances in Neural Information Processing Systems},
 editor = {A. Oh and T. Naumann and A. Globerson and K. Saenko and M. Hardt and S. Levine},
 pages = {28072--28090},
 publisher = {Curran Associates, Inc.},
 title = {Emergent and Predictable Memorization in Large Language Models},
 url = {https://proceedings.neurips.cc/paper_files/paper/2023/file/59404fb89d6194641c69ae99ecdf8f6d-Paper-Conference.pdf},
 volume = {36},
 year = {2023}
}

@article{hupkes2023,
author = {Hupkes, Dieuwke and Giulianelli, Mario and Dankers, Verna and Artetxe, Mikel and Elazar, Yanai and Pimentel, Tiago and Christodoulopoulos, Christos and Lasri, Karim and Saphra, Naomi and Sinclair, Arabella and Ulmer, Dennis and Schottmann, Florian and Batsuren, Khuyagbaatar and Sun, Kaiser and Sinha, Koustuv and Khalatbari, Leila and Ryskina, Maria and Frieske, Rita and Cotterell, Ryan and Jin, Zhijing},
year = {2023},
month = {10},
pages = {1161-1174},
title = {A taxonomy and review of generalization research in NLP},
volume = {5},
journal = {Nature Machine Intelligence},
doi = {10.1038/s42256-023-00729-y}
}

@inproceedings{dankers-etal-2021-generalising,
    title = "Generalising to {G}erman Plural Noun Classes, from the Perspective of a Recurrent Neural Network",
    author = "Dankers, Verna  and
      Langedijk, Anna  and
      McCurdy, Kate  and
      Williams, Adina  and
      Hupkes, Dieuwke",
    editor = "Bisazza, Arianna  and
      Abend, Omri",
    booktitle = "Proceedings of the 25th Conference on Computational Natural Language Learning",
    month = nov,
    year = "2021",
    address = "Online",
    publisher = "Association for Computational Linguistics",
    url = "https://aclanthology.org/2021.conll-1.8",
    doi = "10.18653/v1/2021.conll-1.8",
    pages = "94--108",
}

@article{levshina2023gradientorder,
url = {https://doi.org/10.1515/ling-2021-0098},
title = {Why we need a gradient approach to word order},
title = {},
author = {Natalia Levshina and Savithry Namboodiripad and Marc Allassonnière-Tang and Mathew Kramer and Luigi Talamo and Annemarie Verkerk and Sasha Wilmoth and Gabriela Garrido Rodriguez and Timothy Michael Gupton and Evan Kidd and Zoey Liu and Chiara Naccarato and Rachel Nordlinger and Anastasia Panova and Natalia Stoynova},
pages = {825--883},
volume = {61},
number = {4},
journal = {Linguistics},
doi = {doi:10.1515/ling-2021-0098},
year = {2023},
lastchecked = {2024-06-17}
}

@misc{merrill2024evaluatingngramnoveltylanguage,
      title={Evaluating $n$-Gram Novelty of Language Models Using Rusty-DAWG}, 
      author={William Merrill and Noah A. Smith and Yanai Elazar},
      year={2024},
      eprint={2406.13069},
      archivePrefix={arXiv},
      primaryClass={cs.CL},
      url={https://arxiv.org/abs/2406.13069}, 
}

@inproceedings{baroni2022,
    title = "On the proper role of linguistically-oriented deep net analysis in linguistic theorizing",
    author = "Marco Baroni",
    editor = "Shalom Lappin",
    booktitle = "Algebraic systems and the representation of linguistic knowledge",
    year = "2022",
    publisher = "CRC Press",
    pages = "5--22",
}

@inproceedings{wilson-frank-2023-inductive,
    title = "Inductive Bias Is in the Eye of the Beholder",
    author = "Wilson, Michael  and
      Frank, Robert",
    editor = "Hupkes, Dieuwke  and
      Dankers, Verna  and
      Batsuren, Khuyagbaatar  and
      Sinha, Koustuv  and
      Kazemnejad, Amirhossein  and
      Christodoulopoulos, Christos  and
      Cotterell, Ryan  and
      Bruni, Elia",
    booktitle = "Proceedings of the 1st GenBench Workshop on (Benchmarking) Generalisation in NLP",
    month = dec,
    year = "2023",
    address = "Singapore",
    publisher = "Association for Computational Linguistics",
    url = "https://aclanthology.org/2023.genbench-1.12",
    doi = "10.18653/v1/2023.genbench-1.12",
    pages = "152--162",
}

@inproceedings{oh-etal-2024-frequency,
    title = "Frequency Explains the Inverse Correlation of Large Language Models{'} Size, Training Data Amount, and Surprisal{'}s Fit to Reading Times",
    author = "Oh, Byung-Doh  and
      Yue, Shisen  and
      Schuler, William",
    editor = "Graham, Yvette  and
      Purver, Matthew",
    booktitle = "Proceedings of the 18th Conference of the European Chapter of the Association for Computational Linguistics (Volume 1: Long Papers)",
    month = mar,
    year = "2024",
    address = "St. Julian{'}s, Malta",
    publisher = "Association for Computational Linguistics",
    url = "https://aclanthology.org/2024.eacl-long.162",
    pages = "2644--2663",
}

@inproceedings{huebner-etal-2021-babyberta,
    title = "{B}aby{BERT}a: Learning More Grammar With Small-Scale Child-Directed Language",
    author = "Huebner, Philip A.  and
      Sulem, Elior  and
      Cynthia, Fisher  and
      Roth, Dan",
    editor = "Bisazza, Arianna  and
      Abend, Omri",
    booktitle = "Proceedings of the 25th Conference on Computational Natural Language Learning",
    month = nov,
    year = "2021",
    address = "Online",
    publisher = "Association for Computational Linguistics",
    url = "https://aclanthology.org/2021.conll-1.49",
    doi = "10.18653/v1/2021.conll-1.49",
    pages = "624--646",
}

@misc{prashanth2024recitereconstructrecollectmemorization,
      title={Recite, Reconstruct, Recollect: Memorization in LMs as a Multifaceted Phenomenon}, 
      author={USVSN Sai Prashanth and Alvin Deng and Kyle O'Brien and Jyothir S V au2 and Mohammad Aflah Khan and Jaydeep Borkar and Christopher A. Choquette-Choo and Jacob Ray Fuehne and Stella Biderman and Tracy Ke and Katherine Lee and Naomi Saphra},
      year={2024},
      eprint={2406.17746},
      archivePrefix={arXiv},
      primaryClass={cs.CL},
      url={https://arxiv.org/abs/2406.17746}, 
}

@inproceedings{van-der-wal-etal-2022-birth,
    title = "The Birth of Bias: A case study on the evolution of gender bias in an {E}nglish language model",
    author = "Van Der Wal, Oskar  and
      Jumelet, Jaap  and
      Schulz, Katrin  and
      Zuidema, Willem",
    editor = "Hardmeier, Christian  and
      Basta, Christine  and
      Costa-juss{\`a}, Marta R.  and
      Stanovsky, Gabriel  and
      Gonen, Hila",
    booktitle = "Proceedings of the 4th Workshop on Gender Bias in Natural Language Processing (GeBNLP)",
    month = jul,
    year = "2022",
    address = "Seattle, Washington",
    publisher = "Association for Computational Linguistics",
    url = "https://aclanthology.org/2022.gebnlp-1.8",
    doi = "10.18653/v1/2022.gebnlp-1.8",
    pages = "75--75",
}

@inproceedings{franke2019subjectivity,
  title={Subjectivity-based adjective ordering maximizes communicative success.},
  author={Franke, Michael and Scontras, Gregory and Simonic, Mihael},
  booktitle={CogSci},
  pages={344--350},
  year={2019}
}

@article{scontras2019,
author = {Scontras, Gregory and Degen, Judith and Goodman, Noah},
year = {2019},
month = {11},
pages = {1-21},
title = {On the grammatical source of adjective ordering preferences},
volume = {12},
journal = {Semantics and Pragmatics},
doi = {10.3765/sp.12.7}
}

@article{ARNON201067,
title = {More than words: Frequency effects for multi-word phrases},
journal = {Journal of Memory and Language},
volume = {62},
number = {1},
pages = {67-82},
year = {2010},
issn = {0749-596X},
doi = {https://doi.org/10.1016/j.jml.2009.09.005},
url = {https://www.sciencedirect.com/science/article/pii/S0749596X09000965},
author = {Inbal Arnon and Neal Snider}
}

@inproceedings{abdou-etal-2022-word,
    title = "Word Order Does Matter and Shuffled Language Models Know It",
    author = "Abdou, Mostafa  and
      Ravishankar, Vinit  and
      Kulmizev, Artur  and
      S{\o}gaard, Anders",
    editor = "Muresan, Smaranda  and
      Nakov, Preslav  and
      Villavicencio, Aline",
    booktitle = "Proceedings of the 60th Annual Meeting of the Association for Computational Linguistics (Volume 1: Long Papers)",
    month = may,
    year = "2022",
    address = "Dublin, Ireland",
    publisher = "Association for Computational Linguistics",
    url = "https://aclanthology.org/2022.acl-long.476",
    doi = "10.18653/v1/2022.acl-long.476",
    pages = "6907--6919",
}

@article{MARTIN1969471,
title = {Some competence-process relationships in noun phrases with prenominal and postnominal adjectives},
journal = {Journal of Verbal Learning and Verbal Behavior},
volume = {8},
number = {4},
pages = {471-480},
year = {1969},
issn = {0022-5371},
doi = {https://doi.org/10.1016/S0022-5371(69)80091-5},
url = {https://www.sciencedirect.com/science/article/pii/S0022537169800915},
author = {J.E. Martin}
}

@inproceedings{sinha-etal-2021-masked,
    title = "Masked Language Modeling and the Distributional Hypothesis: Order Word Matters Pre-training for Little",
    author = "Sinha, Koustuv  and
      Jia, Robin  and
      Hupkes, Dieuwke  and
      Pineau, Joelle  and
      Williams, Adina  and
      Kiela, Douwe",
    editor = "Moens, Marie-Francine  and
      Huang, Xuanjing  and
      Specia, Lucia  and
      Yih, Scott Wen-tau",
    booktitle = "Proceedings of the 2021 Conference on Empirical Methods in Natural Language Processing",
    month = nov,
    year = "2021",
    address = "Online and Punta Cana, Dominican Republic",
    publisher = "Association for Computational Linguistics",
    url = "https://aclanthology.org/2021.emnlp-main.230",
    doi = "10.18653/v1/2021.emnlp-main.230",
    pages = "2888--2913",
}

@book{Dixon1982,
  address   = {Berlin},
  author    = {Dixon, Robert M. W.},
  publisher = {Mouton de Gruyter},
  title     = {‘Where Have All the Adjectives Gone?’ and other Essays in Semantics and Syntax},
  year      = {1982}
}

@book{cinque2010syntax,
    author = {Cinque, Guglielmo},
    title = "{The Syntax of Adjectives: A Comparative Study}",
    publisher = {The MIT Press},
    year = {2010},
    month = {07},
    isbn = {9780262014168},
    doi = {10.7551/mitpress/9780262014168.001.0001},
    url = {https://doi.org/10.7551/mitpress/9780262014168.001.0001},
}

@book{sweet1898new,
  title={A New English Grammar, Logical and Historical},
  author={Sweet, H.},
  number={pt. 2},
  series={A New English Grammar, Logical and Historical},
  url={https://books.google.nl/books?id=U60VAAAAYAAJ},
  year={1898},
  publisher={Clarendon Press}
}

@article{Vecchi2017SpicyAA,
  title={Spicy Adjectives and Nominal Donkeys: Capturing Semantic Deviance Using Compositionality in Distributional Spaces},
  author={Eva Maria Vecchi and Marco Marelli and Roberto Zamparelli and Marco Baroni},
  journal={Cognitive science},
  year={2017},
  volume={41 1},
  pages={
          102-136
        },
  url={https://api.semanticscholar.org/CorpusID:205031527}
}

@inproceedings{Vecchi2013StudyingTR,
  title={Studying the Recursive Behaviour of Adjectival Modification with Compositional Distributional Semantics},
  author={Eva Maria Vecchi and Roberto Zamparelli and Marco Baroni},
  booktitle={Conference on Empirical Methods in Natural Language Processing},
  year={2013},
  url={https://api.semanticscholar.org/CorpusID:10663318}
}

@Inbook{Sproat1991,
author="Sproat, Richard
and Shih, Chilin",
editor="Georgopoulos, Carol
and Ishihara, Roberta",
title="The Cross-Linguistic Distribution of Adjective Ordering Restrictions",
bookTitle="Interdisciplinary Approaches to Language: Essays in Honor of S.-Y. Kuroda",
year="1991",
publisher="Springer Netherlands",
address="Dordrecht",
pages="565--593",
isbn="978-94-011-3818-5",
doi="10.1007/978-94-011-3818-5_30",
url="https://doi.org/10.1007/978-94-011-3818-5_30"
}

@inbook{Cinque_1996, 
place={Cambridge}, series={Cambridge Studies in Linguistics}, title={On the evidence for partial N-movement in the Romance DP}, booktitle={Italian Syntax and Universal Grammar}, publisher={Cambridge University Press}, author={Cinque, Guglielmo}, year={1996}, pages={287–309}, collection={Cambridge Studies in Linguistics}
}

@article{truswell2009,
 ISSN = {00243892, 15309150},
 URL = {http://www.jstor.org/stable/40284329},
 author = {Robert Truswell},
 journal = {Linguistic Inquiry},
 number = {3},
 pages = {525--533},
 publisher = {The MIT Press},
 title = {Attributive Adjectives and Nominal Templates},
 urldate = {2024-06-07},
 volume = {40},
 year = {2009}
}

@inbook{scott2022,
author = {Scott, Gary-John},
year = {2002},
month = {10},
pages = {91-120},
title = {Stacked Adjectival Modification and the Structure of Nominal Phrases},
isbn = {9780195148794},
doi = {10.1093/oso/9780195148794.003.0004}
}

@inproceedings{malkin-etal-2021-studying,
    title = "Studying word order through iterative shuffling",
    author = "Malkin, Nikolay  and
      Lanka, Sameera  and
      Goel, Pranav  and
      Jojic, Nebojsa",
    editor = "Moens, Marie-Francine  and
      Huang, Xuanjing  and
      Specia, Lucia  and
      Yih, Scott Wen-tau",
    booktitle = "Proceedings of the 2021 Conference on Empirical Methods in Natural Language Processing",
    month = nov,
    year = "2021",
    address = "Online and Punta Cana, Dominican Republic",
    publisher = "Association for Computational Linguistics",
    url = "https://aclanthology.org/2021.emnlp-main.809",
    doi = "10.18653/v1/2021.emnlp-main.809",
    pages = "10351--10366",
}

@misc{lesci2024causal,
      title={Causal Estimation of Memorisation Profiles}, 
      author={Pietro Lesci and Clara Meister and Thomas Hofmann and Andreas Vlachos and Tiago Pimentel},
      year={2024},
      eprint={2406.04327},
      archivePrefix={arXiv},
      primaryClass={cs.LG}
}

@article{TrotzkeWittenberg,
    url = {https://doi.org/10.1515/ling-2019-0001},
    title = {Long-standing issues in adjective order and corpus evidence for a multifactorial approach},
    title = {},
    author = {Andreas Trotzke and Eva Wittenberg},
    pages = {273--282},
    volume = {57},
    number = {2},
    journal = {Linguistics},
    doi = {doi:10.1515/ling-2019-0001},
    year = {2019},
    lastchecked = {2024-06-09}
}

@inproceedings{futrell-levy-2019-rnns,
    title = "Do {RNN}s learn human-like abstract word order preferences?",
    author = "Futrell, Richard  and
      Levy, Roger P.",
    editor = "Jarosz, Gaja  and
      Nelson, Max  and
      O{'}Connor, Brendan  and
      Pater, Joe",
    booktitle = "Proceedings of the Society for Computation in Linguistics ({SC}i{L}) 2019",
    year = "2019",
    url = "https://aclanthology.org/W19-0106",
    doi = "10.7275/jb34-9986",
    pages = "50--59",
}

@book{vendler1968adjectives,
  title={Adjectives and Nominalizations},
  author={Vendler, Z.},
  isbn={9789027900838},
  lccn={67030544},
  series={Papers on formal linguistics},
  url={https://books.google.nl/books?id=j620AAAAIAAJ},
  year={1968},
  publisher={Mouton}
}

@article{DBLP:journals/corr/abs-2309-07311,
  author       = {Angelica Chen and
                  Ravid Schwartz{-}Ziv and
                  Kyunghyun Cho and
                  Matthew L. Leavitt and
                  Naomi Saphra},
  title        = {Sudden Drops in the Loss: Syntax Acquisition, Phase Transitions, and
                  Simplicity Bias in MLMs},
  journal      = {CoRR},
  volume       = {abs/2309.07311},
  year         = {2023},
  url          = {https://doi.org/10.48550/arXiv.2309.07311},
  doi          = {10.48550/ARXIV.2309.07311},
  eprinttype    = {arXiv},
  eprint       = {2309.07311},
  bibsource    = {dblp computer science bibliography, https://dblp.org}
}
\bibliographystyle{acl_natbib}

\clearpage
\appendix
\section{Complementary Results}
\subsection{CAP Examples}\label{app:cap-examples}
We provide a sample of CAP examples in Table~\ref{tab:cap_examples}, categorised by the contextual and isolated AOP scores of Pythia-12b.

\begin{table*}[t]
    \centering
    \renewcommand{\arraystretch}{1.1}
    \begin{tabular}{@{}r p{4.5in} c c@{}} 
&\textsf{CAP Sentence} & AOP$(\bullet)$ & AOP$(\bullet|c)$ \\[2pt]\arrayrulecolor{black}\hline
\multicolumn{4}{c}{\cellcolor{lg}\textcolor{white}{\textsf{CAP order always preferred}}}\\
1. & \textit{The seat comprises the \textbf{following} \textbf{electoral} wards} & $14.5$ & $23.6$\\
2. & \textit{The local wildlife includes leopards, deer, monkeys, several kinds of pheasants, the marbled polecat and the increasingly \textbf{rare} \textbf{flying} squirrel} & $11.1$ & $24.9$\\
3. & \textit{He was elected as leader of the Labour Party in the first round of voting using the \textbf{single} \textbf{transferable} vote} & $13.4$ & $21.7$\\
4. & \textit{Hoggett and mutton can taste more flavorful than lamb because they contain a higher concentration of species- \textbf{characteristic} \textbf{fatty} acids} & $12.8$ & $22.1$\\
5. & \textit{Jimmy seemed ready to take up that challenge, and entered upon an argument calculated to prove that he was a \textbf{mild} \textbf{mannered} individual} & $14.4$ & $18.4$\\
\multicolumn{4}{c}{\cellcolor{dr}\textcolor{white}{\textsf{Swapped order always preferred}}}\\
6. & \textit{Mm, well, the extraction of \textbf{foreign} \textbf{rectal} bodies} & $-11.3$ & $-11.3$\\
7. & \textit{did they sing it's an \textbf{old} \textbf{grand} flag} & $-7.7$ & $-14.1$\\
8. & \textit{His great jaw protruded frightfully, with the \textbf{upper} \textbf{thin} lip} & $-6.6$ & $-6.4$\\
9. & \textit{And Stuart sat down in, because they got these \textbf{wooden} \textbf{old} chairs} & $-6.9$ & $-6.0$\\
10. & \textit{It does this by following the \textbf{grammatical} \textbf{basic} rules} & $-5.9$ & $-6.1$\\
\multicolumn{4}{c}{\cellcolor{dg}\textcolor{white}{\textsf{Context improves AOP}}}\\
11. & \textit{Be it thy course to \textbf{busy} \textbf{giddy} minds} & $1.9$ & $26.2$\\
12. & \textit{I have not passed through fire and death to \textbf{bandy} \textbf{crooked} words} & $5.3$ & $29.2$\\
13. & \textit{"I wish to preach, not the doctrine of ignoble ease, but the doctrine of the strenuous life; the life of toil and effort; of labor and strife; to preach that highest form of success which comes, not to the man who desires \textbf{mere} \textbf{easy} peace} & $2.3$ & $21.8$\\
14. & \textit{The "Yonsei" are the subject of on- \textbf{going} \textbf{academic} research} & $-0.2$ & $18.1$\\
15. & \textit{The city flag of Portland, Oregon, consists of a green field on which is placed a white 4-pointed star from which radiate blue stripes, each bordered by L- \textbf{shaped} \textbf{yellow} lines} & $-1.0$ & $16.9$\\
\multicolumn{4}{c}{\cellcolor{lr}\textcolor{white}{\textsf{Context worsens AOP}}}\\
16. & \textit{One expected that it earned a \textbf{alone} \textbf{decisive} battle} & $9.1$ & $-1.1$\\
17. & \textit{Meager is known for its \textbf{last} \textbf{major} eruption} & $12.8$ & $4.8$\\
18. & \textit{As a matter of fact, no body in the universe revolves about the \textbf{exact} \textbf{geometrical} center} & $7.6$ & $-0.4$\\
19. & \textit{E \textbf{major} is a \textbf{musical} \textbf{major} scale} & $4.8$ & $-2.8$\\
20. & \textit{That worthy was seated before a table spread with papers, and as Tom entered or rather was pushed into his presence he compressed his \textbf{beetling} \textbf{black} brows} & $5.1$ & $-2.5$\\
\end{tabular}
    \caption{Most salient examples of the four quadrants in Figure~\ref{fig:role_of_context}.}
    \label{tab:cap_examples}
\end{table*}

\subsection{Overlap and Correlations to Cognitive Predictors}\label{app:cognitive-predictors}
Overlap between various cognitive predictors and AOP-$\Delta$ scores of Pythia-1.4b::
\begin{center}
\includegraphics[height=1.43in,clip,trim={0.25cm 0 0.25cm 0}]{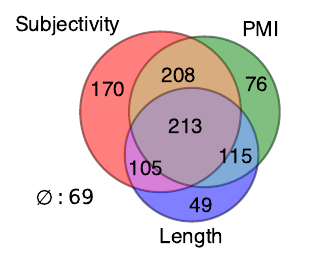}\quad
\includegraphics[height=1.43in,clip,trim={0.25cm 0 0.25cm 0}]{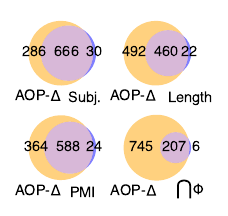}
\end{center}\vspace{1cm}

Correlations during training of \aopDelta scores (Pythia-1.4b), with scores derived from various cognitive predictors:
\includegraphics[width=\columnwidth]{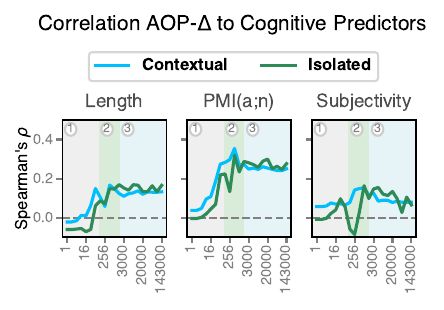}

\subsection{Relative Number of Bigram Occurrences}\label{app:split-sizes}
\includegraphics[width=\columnwidth]{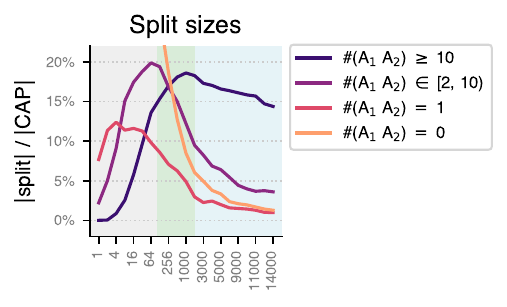}
Relative sizes of the 4 splits of CAP based on the number of bigram occurrences seen at each point in training.
Note that an additional constraint on each split is that the swapped adjective order has not been seen at all, which explains why the 4 splits do not sum up to 1.

\subsection{One-shot AOP}\label{app:one-shot-aop}
\begin{figure*}[t]
    \centering
    \includegraphics[width=\textwidth]{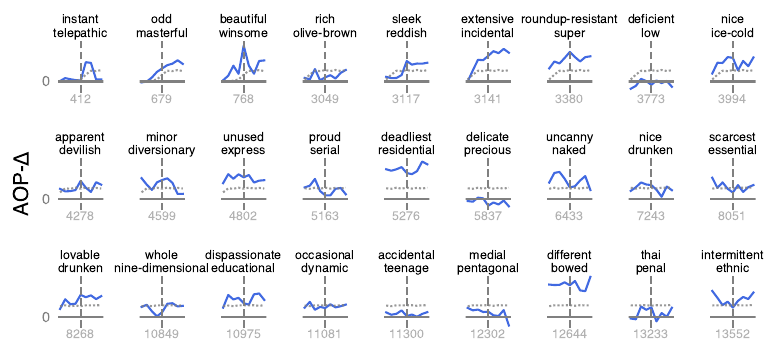}
    \caption{
        The AOP-$\Delta$ scores (Pythia-1.4b) across time for the subset of adjective pairs that are seen exactly once up to the 14,000\textsuperscript{th} batch.
        Each curve is centered around the point where the adjective pair was encountered.
        The dotted line denotes the average AOP-$\Delta$ across all CAP items.
    }
    \label{fig:one-shot-aop}
\end{figure*}

In Figure~\ref{fig:one-shot-aop} we plot the AOP-$\Delta$ scores for the 27 adjective pairs that are seen exactly once in the first 10\% of training.


\subsection{Context Improvement Ratios}\label{sec:app_context_ratios}
\begin{center}
\includegraphics[width=0.6\columnwidth]{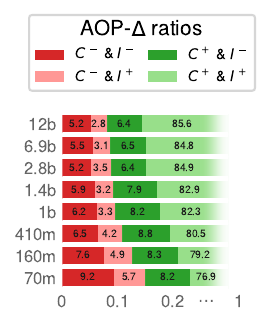}\\    
\end{center}
The relative number of items that are affected by context with respect to isolated AOP-$\Delta$ scores for each model size. 
These ratios correspond to the four quadrants of Figure~\ref{fig:role_of_context}A.


\subsection{Context Length Impact}\label{sec:random_context_len}
\begin{center}
\includegraphics[width=2in]{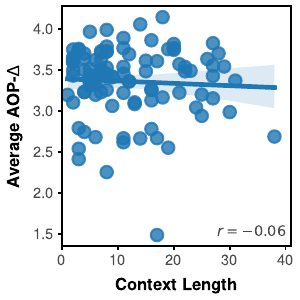}\\    
\end{center}
Relation between context length and contextual AOP-$\Delta$ scores, averaged over \textit{randomly sampled} contexts. The context length effect shown in Figure~\ref{fig:len_entropy} disappears when averaging over random contexts, showing that contextual gain is not driven by context length itself.

\end{document}